\documentclass[manuscript,screen,review]{acmart}

\AtBeginDocument{%
  \providecommand\BibTeX{{%
    \normalfont B\kern-0.5em{\scshape i\kern-0.25em b}\kern-0.8em\TeX}}}

\setcopyright{acmcopyright}
\copyrightyear{2018}
\acmYear{2018}
\acmDOI{XXXXXXX.XXXXXXX}

\acmConference[Conference acronym 'XX]{Make sure to enter the correct
  conference title from your rights confirmation emai}{June 03--05,
  2018}{Woodstock, NY}
\acmPrice{15.00}
\acmISBN{978-1-4503-XXXX-X/18/06}

\usepackage{booktabs} 
\usepackage{makecell}

\begin{document}

\title{Enhanced Infield Agriculture with Interpretable Machine Learning Approaches for Crop Classification}





\author{Sudi Murindanyi}
\email{sudi.murindanyi@students.mak.ac.ug}
\orcid{0009-0001-3667-5814}
\affiliation{%
 \institution{Makerere University}
 \country{Uganda}}

\author{Joyce Nakatumba-Nabende}
\email{joyce.nabende@mak.ac.ug}
\orcid{0000-0002-0108-3798}
\affiliation{%
  \institution{Makerere University}
  \country{Uganda}}

\author{Rahman Sanya}
\email{hbasanya@gmail.com}
\orcid{0000-0002-4631-3980}
\affiliation{%
  \institution{Makerere University}
  \country{Uganda}
  }

\author{Rose Nakibuule}
\email{rnakibuule@gmail.com}
\orcid{0009-0009-1838-9214}
\affiliation{%
  \institution{Makerere University}
  \country{Uganda}
  }


\author{Andrew Katumba}
\email{andrew.katumba@mak.ac.ug}
\orcid{0000-0001-7699-8426}
\affiliation{%
  \institution{Makerere University}
  \country{Uganda}
  }





\begin{abstract}
The increasing popularity of Artificial Intelligence (AI) in recent years has led to a surge in interest in image classification, especially in the agricultural sector. With the help of Computer Vision (CV), Machine Learning (ML), and Deep Learning (DL), the sector has undergone a significant transformation, leading to the development of new techniques for crop classification in the field. Despite the extensive research on various image classification techniques, most have limitations such as low accuracy, limited use of data, and a lack of reporting model size and prediction. The most significant limitation of all is the need for model explainability. This research evaluates four different approaches for crop classification, namely traditional ML with handcrafted feature extraction methods like Scale-Invariant Feature Transform (SIFT), Oriented FAST and Rotated BRIEF (ORB), and Color Histogram; Custom Designed Convolution Neural Network (CNN) and established DL architecture like AlexNet; transfer learning on five models pre-trained using ImageNet such as EfficientNetV2, ResNet152V2, Xception, Inception-ResNetV2, MobileNetV3; and cutting-edge foundation models like YOLOv8 (You Only Look Once) and DINOv2, a self-supervised Vision Transformer Model. All models performed well, but Xception outperformed all of them in terms of generalization, achieving 98\% accuracy on the test data, with a model size of 80.03 MB and a prediction time of 0.0633 seconds. A key aspect of this research was the application of Explainable AI (XAI) to provide the explainability of all the models. This journal presents the explainability of Xception model with LIME (Local Interpretable Model-agnostic Explanations), SHAP (SHapley Additive exPlanations), and GradCAM (Gradient-weighted Class Activation Mapping), ensuring transparency and trustworthiness in the models' predictions. This study highlights the importance of selecting the right model according to task-specific needs. It also underscores the important role of explainability in deploying AI in agriculture, providing insightful information to help enhance AI-driven crop management strategies, leading to more efficient and sustainable farming practices.
\end{abstract}

\ccsdesc[500]{Do Not Use This Code~Generate the Correct Terms for Your Paper}
\ccsdesc[300]{Do Not Use This Code~Generate the Correct Terms for Your Paper}
\ccsdesc{Do Not Use This Code~Generate the Correct Terms for Your Paper}
\ccsdesc[100]{Do Not Use This Code~Generate the Correct Terms for Your Paper}

\keywords{Do, Not, Us, This, Code, Put, the, Correct, Terms, for,
  Your, Paper}

\received{20 February 2007}
\received[revised]{12 March 2009}
\received[accepted]{5 June 2009}

\maketitle

\section{Introduction}
\subsection{Oveview}
Image classification is regarded as one of the major topics in computer vision and artificial intelligence (AI) \cite{singh2020image}\cite{peng2022survey}. It involves correctly identifying and categorising objects in images. Various techniques, such as handcrafted feature extraction methods, are used to extract features from images and then are used for image classification using machine learning classification algorithms \cite{kaur2019various}. Moreover, deep learning (DL), a powerful subset of machine learning (ML), has currently achieved impressive results in computer vision applications such as image classification, object detection, and image processing \cite{aditi2021review}. In the deep learning approach, feature extraction and classification are seamlessly integrated simultaneously to classify images. In addition to the methods mentioned above, transfer learning in deep learning has become a crucial strategy, particularly when having a small amount of labelled data \cite{ali2023transfer}. It involves leveraging knowledge from one problem domain to solve a different but related problem. For instance, a model trained on Large-scale datasets like ImageNet can be adapted to classify images in specialized domains, reducing the need for extensive labelling \cite{ali2023transfer}. Furthermore, foundation models, large-scale pre-trained models trained on diverse datasets, have revolutionized the field \cite{yang2023foundation}\cite{meng2023foundation}\cite{liu2023large}. By fine-tuning these models to specific tasks, significant performance improvements can be achieved even with limited labelled data. This method utilizes the already learned complex and profound representations of foundational models, resulting in a more efficient and effective knowledge transfer to novel tasks \cite{yang2023foundation}. The challenge is that most researchers only consider one or two of these approaches to develop a classification model for their task. It should be noted that no conclusion should be made before evaluating the options. Each of them has the potential to perform well if used with better approaches and techniques. The best practice would be to evaluate them and choose the most suitable one for the task. Additionally, It is important to ensure that all machine learning models are explainable. Despite their reputation as black boxes, these models offer more than just impressive accuracy and favourable performance metrics \cite{zhou2021evaluating}. To fully utilize the capabilities of these models, it is crucial to explain how they make decisions. By creating visual representations demonstrating how models arrive at their conclusions. With this, individuals from different fields can better understand the results more intuitively. Therefore, Developing and presenting interpretable outcomes is necessary for promoting trust in AI technology.

The agricultural sector has experienced a significant transformation in recent years, mainly due to the resurgence of AI \cite{Vasavi2022}. Thanks to Computer Vision(CV), ML, and DL advancements, sophisticated algorithms now power intelligent systems that automate various agricultural processes \cite{Saleem2021}. One specific area that still needs more significant transformation is the classification of plants in the infield. Farmers can gain precise insights into this crucial area with modern AI-driven crop classification. Targeting this area can enhance precision agriculture by providing a detailed understanding of crop variations and infield needs with the help of modern AI. Additionally, this can facilitate crop yield estimation, which is essential in ensuring optimal agricultural output and sustainability \cite{patil2019machine}.

This research targeted to develop and assess the four distinct image classification methodologies within the agricultural domain. The initial approach involved manually extracting features using handcrafted feature extraction techniques from images and using traditional machine learning algorithms for classification. Subsequently, we explored the potential of deep learning architectures: firstly, a custom convolutional neural network (CNN) model was designed from the ground up to perform classification task, and secondly, a well-established CNN architecture, AlexNet architecture, was used for classification. The third approach involved applying transfer learning to preeminent architectures trained using Imagenet, an effective strategy when dealing with limited labelled data. Finally, extensively utilized pre-trained foundation models, known to be trained on vast and varied datasets, to develop a crop classifier. Using agricultural imagery, these classifiers were trained and evaluated on the data collected in the infield and different types of crops were considered. Finally, crucial to the study was the integration of Explainable AI (XAI) tools, such as Gradient-weighted Class Activation Mapping (Grad-CAM), to ensure the transparency and interpretability of models. The goal was to point out the decision-making process of AI models by providing easy-to-understand visual explanations for their predictions. This approach helps to increase trust in the model's output while offering people from other fields who lack AI expertise the ability to understand and implement the results in agricultural practices. This can bridges the gap between advanced AI applications and practical agricultural implementation.

\subsection{Background}

\subsubsection{\textbf{Classification Using Handcrafted Features and Classical Machine Learning Models:}}
Image classification heavily relies on the extracted features \cite{abdelmoneim2019comparative}\cite{kaur2019various}. These features can be classified based on colour, shape, or texture \cite{kaur2019various}. Different handcrafted feature extraction methods can be used to extract these features, which can then be used to classify crops with traditional ML models \cite{song2019object}. One of the main challenges using this approach is the quality and relevance of extracted features. The performance of these models is directly linked to how well these features represent the underlying patterns in the images, and sometimes, it takes much work to do it. Various handcrafted feature extraction techniques have been employed for image classification \cite{gupta2019improved}\cite{kaur2019various}, including Scale-Invariant Feature Transform (SIFT), Oriented FAST and Rotated BRIEF (ORB), Speeded-Up Robust Features (SURT), and colour histograms, to mention a few. Only three were implemented for this research: SIFT, ORB, and colour histogram. SIFT, a widely recognized approach, detects critical points in an image that remain invariant under affine, rotational, and scale transformations. This technique can represent leaves' texture, shape, and appearance, proving robust against variations in lighting and viewpoint \cite{guo2018research}\cite{gupta2019improved}. On the other hand, ORB is a feature extraction algorithm that detects essential areas in images using corner points or regions with significant intensity shifts. It employs a pyramid scheme with a FAST keypoint detector and a BRIEF keypoint descriptor \cite{zhang2018improved}. The colour histogram, however, represents the frequency distribution of colours that appear in an image. It is especially effective in distinguishing images based on their unique colour profiles and analyzing the intensity and distribution of colours present in them \cite{mali2014color}. 

While handcrafted feature extraction methods have succeeded in classification tasks, it is essential to acknowledge their limitations \cite{mwadulo2016review}. They rely on domain expertise and manual tuning. In cases where specific leaf textures for crop classification or structural attributes are crucial in distinguishing between crops, they can be effective. However, in cases where the visual patterns and attributes that differentiate crops are complex, high-dimensional, or not easily captured by predefined rules, they may require additional assistance.

\begin{figure}[!htb]
\centering
\includegraphics[width=1\textwidth]{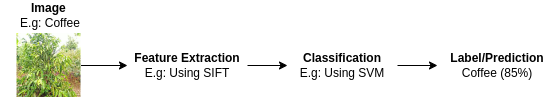}
\caption{Classification using Machine Learning}
\label{fig:ml1}
\end{figure}

After feature extraction, different methods can be used to select the best features, such as Principal Component Analysis (PCA). Then, these features can be used to train any classical ML model, like a Support Vector Machine (SVM), to classify images into different classes. Figure \ref{fig:ml1} shows the visual representation of the process using classical ML models and handcrafted features from image input to prediction of the class. 

\subsubsection {\textbf{Classification Using Deep Learning:}}
Deep learning methods, especially Convolutional Neural Networks (CNNs), have demonstrated remarkable potential in automatically learning and extracting distinctive features directly from image data \cite{ueki2015multi}. CNNs are highly effective in capturing hierarchical representations of images, which allows them to learn complex and abstract patterns that handcrafted features may not explicitly define \cite{ravanbakhsh2015action}\cite{ueki2015multi}. Using large datasets, CNNs can generalize to diverse visual variations and handle complex, high-dimensional feature spaces. This makes them a valuable tool for image recognition and classification tasks.

\begin{figure}[!htb]
\centering
\includegraphics[width=1\textwidth]{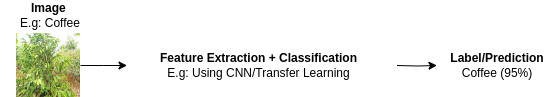}
\caption{Classification using Deep Learning}
\label{fig:ml2}
\end{figure}

Deep learning proves particularly effective when the features differentiating images or highly complex visual patterns are poorly understood. To enhance the accuracy of deep learning models in such contexts, researchers have increasingly turned to CNNs and transfer learning strategies \cite{burri2023exploring}. CNNs, known for their ability to extract spatial hierarchies in image data, can be coupled with transfer learning, allowing pre-trained models on extensive datasets to be fine-tuned for specific tasks \cite{burri2023exploring}\cite{rajpura2017transfer}. These techniques can be combined to solve sophisticated tasks. This approach is invaluable in domains with limited data availability. For this research, a custom CNN model was explicitly designed for crop classification. Also, the AlexNet architecture was trained for comparison purposes, as it is known for its effectiveness in image recognition \cite{sridhar2022classification}. Additionally, transfer learning was done on models pre-trained on ImageNet, including EfficientNetV2 (known for balancing efficiency and accuracy) \cite{tan2021efficientnetv2}, ResNet152V2 (notable for its deep residual learning framework) \cite{ibrahim2020soft}, Xception (which utilizes depthwise separable convolutions) \cite{chollet2017xception}, Inception-ResNetV2 (a hybrid network combining Inception modules with residual connections) \cite{zhang2018residual}, and MobileNetV3 (an optimized version of MobileNet with enhanced performance) \cite{howard2019searching}\cite{qian2021mobilenetv3}. These models were assessed with their distinct architectures and learning capabilities to determine their suitability and effectiveness in classifying crops and weeds in the infield. Figure \ref{fig:ml2} shows the visual representation of the process using deep learning models from image input to class prediction.

\subsubsection {\textbf{Crop Classification Using Foundation Models:}}
Foundation models represent a distinct class of Deep Learning models, distinguished by their training on immense amounts of data and adaptability to various downstream tasks \cite{yang2023foundation}. Unlike transfer learning, their comprehensive training allows them to understand patterns and features broadly. These models can be fine-tuned for specific tasks, such as crop classification, using smaller labelled datasets, often leading to impressive results. While foundation models leverage transfer learning, their approach differs from conventional methods. Their ability to be used in many ways allows for new approaches in agriculture, like predicting crop diseases with very small data and yield estimation. However, they come with challenges, one being that the computational demands for fine-tuning can be significant, and their back box nature often complicates the interpretation of predictions, a crucial factor for practical agricultural applications \cite{meng2023foundation}.

Two highly-regarded foundation models were evaluated for the crop classification tasks: YOLOv8 and DINOv2. YOLOv8, the latest version of the YOLO (You Only Look Once) series, is known for its real-time object detection capabilities, which it achieves through an anchor-free detection mechanism, an enriched feature pyramid network, and a recalibrated loss function \cite{terven2023comprehensive}. This unique combination allows superior object detection and classification by predicting bounding boxes and associated class probabilities using a single convolutional neural network. On the other hand, DINOv2 is a self-supervised learning technique that can discern robust visual features without requiring labelled data \cite{oquab2023dinov2}. By leveraging pretraining methodologies and a curated dataset, DINOv2 provides versatile visual features that can operate seamlessly across diverse image distributions and tasks without requiring further fine-tuning \cite{oquab2023dinov2}. Figure \ref{fig:ml1} shows the visual representation of the process using foundation models from image input to class prediction. Some foundation models can directly extract features and perform classification, like YOLOv8, and others can first extract features and perform classification separately, like DINOv2.

\begin{figure}[!htb]
\centering
\includegraphics[width=1\textwidth]{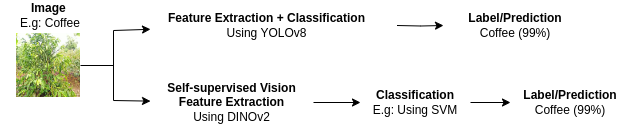}
\caption{Classification Using Foundation Models}
\label{fig:ml3}
\end{figure}

\subsubsection {\textbf{Explainablility of Crop Classification Models:}}
While advanced AI models have improved classification accuracy and other metrics, there is still a need for transparency and interpretability because their results are considered a black box and cannot be understood or trusted by anyone \cite{gastounioti2020time}\cite{tamagnini2017interpreting}. Understanding predictions is crucial for strategic planning, yield estimation, and sustainable agriculture. Delving deeper into the reasons for each classification provides insights into which features the model considers most significant, guiding the understanding of why predictions are made in a certain way \cite{basha2018evaluating}. Insights like these are crucial for making informed decisions and implementing sustainable agricultural practices.

One solution to the challenge of explainability in AI is using eXplainable AI (XAI) tools, which include LIME (Local Interpretable Model-agnostic Explanations) \cite{10.1145/2939672.2939778}, SHAP (SHapley Additive exPlanations) \cite{NIPS2017_8a20a862}, and GradCAM (Gradient-weighted Class Activation Mapping) \cite{Selvaraju_2017_ICCV}, mention but a few. LIME perturbs input data and observes prediction changes for local explanations. SHAP values, rooted in cooperative game theory, allocate contributions of each feature to every possible prediction. GradCAM provides visual explanations from convolutional networks via class-discriminative regions. These techniques enhance model interpretability, building trust and enabling informed crop classification.

\begin{figure}[!htb]
\centering
\includegraphics[width=1\textwidth]{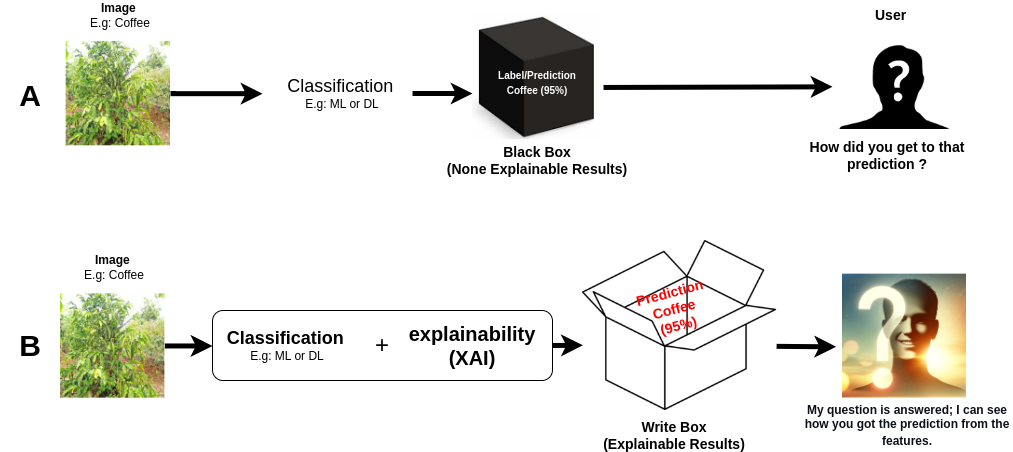}
\caption{Classification Without Explainability Vs Without Explainability}
\label{fig:ml4}
\end{figure}

The remaining sections of this work are organized as follows: Section two provides a systematic literature review of the related work done around some of the methods used for crop classification. Section three presents some of the gaps in reviewed related work. Section four presents the journal contributions that tried to solve the gaps in related work. Section five presents the methodologies used, including step-by-step explanations of the tools employed. Section six provides the results of all the methods used and discussions of the outcomes. Finally, section seven presents the conclusion of the work, including a discussion of future work. Figure \ref{fig:ml4} shows the visual representation of classification without XAI (A) and with XAI (B). The first path (A) shows that if you don't use explainability, you get results that are impossible to interpret, and path (B) shows that if you use explainability, you get results that everyone can understand for the features representation in the images.

\section{Related Work}

The section discusses recent research on crop classification using AI-driven models. The literature review was conducted by searching databases for papers published between 2018 and 2023 using specific keywords. Only papers that met strict criteria were considered, including being in English, employing machine learning techniques, and discussing new methodologies, architectures, or significant accuracy improvements.

S. Roopashree et al. \cite{ROOPASHREE2022111484} highlighted the application of machine learning in classifying medicinal plants. Their model, Herbmodel, demonstrated high accuracy by employing feature extraction techniques and a support vector machine classifier. The study achieved an average accuracy of 96.22\% across a dataset of 2515 samples from 40 species. Jyoti Madake et al. \cite{10119325} highlighted the significance of utilizing image processing and feature extraction techniques. They discovered that the Random Forest classifier combined with SIFT produced promising results. The trained model could accurately detect and classify healthy and wilted plants, achieving an impressive accuracy rate of 85.41\% when SIFT was combined with the Random Forest classifier. This performance was comparable to that of existing models.  M. Waqas and N. Fukushima \cite{9306425} showed that AKAZE, SIFT and SURF, a feature descriptor, outperformed others regarding computational speed and classification accuracy. AKAZE achieved a classification accuracy of 81.56\%, slightly comparable to SIFT's 87.25\% and SURF's 85.73\%. Further, Bansal et al. \cite{bansal2021transfer}  Combined deep learning techniques with traditional handcrafted feature extraction to achieve impressive classification results. Their approach used the VGG19 network and various classifiers, with the Random Forest classifier outperforming other classifiers and methods. The experimental evaluation on the Caltech-101 benchmark dataset achieved an accuracy of 93.73\%. Selvaraj et al. \cite{01} employed a pixel-based classification approach for classifying crops from aerial images, achieving high accuracy rates. They utilized Support Vector Machines (SVMs) and Random Forest (RF) machine learning models and combined features derived from vegetation indices (VIs) and PCA for pixel-based classification. The results showed an overall accuracy of up to 97\%, highlighting this approach's effectiveness for crop classification. While classification accuracy is paramount, understanding the reasons behind such classifications is equally vital. Sunil et al. \cite{sunil2022weed} used RGB image texture features to classify weed and crop species. They compared the classification performance of the Support Vector Machine (SVM) model and the deep learning-based visual group geometry 16 (VGG16) model. Their experiment captured 3792 RGB images of crop and weed samples from the greenhouse. The results showed that the VGG16 model classifier had an average f1-score between 93\% and 97.5\%. Abdelmalek et al. \cite{bouguettaya2022deep} reviewed deep learning techniques for crop classification using UAV-based remote sensing images. They found that these techniques can achieve impressive results. In conclusion, they suggested building an intelligent Internet of Drones (IoD) system that could monitor farmlands to improve crop productivity and reduce the need for human intervention. The researchers also recommended developing lightweight versions of deep learning algorithms that can be implemented on small devices with low computational power. This would help reduce energy consumption and increase the flying time of UAVs, which is a critical issue. M.K Dharani et al. \cite{dharani2021review} used deep learning techniques to analyze methods such as ANN, CNN, RNN-LSTM, and Hybrid networks in their research on crop yield prediction. They found CNN to be more effective than ANN, with an accuracy of 87\%, while RNN-LSTM and Hybrid networks achieved even higher accuracy, around 89-90\%. The study suggests future research focus on applying RNN and Hybrid networks to various crops and real-time datasets, emphasizing the increasing role of artificial intelligence in agriculture. Ayshah Chan et al. \cite{chan2023xai} emphasize early crop classification through eXplainable AI (XAI) methods. Their approach spotlights significant timesteps in crop classification, revealing links to physical crop growth milestones. Additionally, Exploring Self-Attention for Crop-type Classification Explainability by Ivica Obadic et al. \cite{obadic2210exploring} introduced an explainability framework for crop-type classification using transformer encoders. Their results emphasized the importance of attention patterns and phenological events for crop classification. Addressing generalization issues in crop-type mapping, Towards Explainable AI4EO by Adel Abbas et al. \cite{abbas2023towards} proposed an Explainable AI approach. Their method, which focuses on temporal shifts in Satellite Image Time Series, demonstrates improved accuracy, emphasizing the importance of understanding and addressing the inner workings of deep learning models in crop classification. 

\section{Gaps In Reviewed Literature}

These studies and many more reviewed collectively underscore the promise of employing image feature extraction alongside machine learning techniques for crop classification and detection and the flourishing role of deep learning in these areas. However, they also point to the necessity for more in-depth research to create classification models that are both robust and accurate. These enhanced models could significantly contribute to advancing agricultural practices and more effective crop management on a larger scale. Notably, current literature reveals a gap in harnessing the capabilities of foundation models in crop classification. Foundation models, known for their vast scale and versatility in various domains, might offer untapped potential in this field. A notable limitation in existing research is the need for more interpretability in the results produced. Understanding and explaining the decisions made by machine learning models is crucial, particularly in applications impacting critical sectors like agriculture. These studies also reveal some critical limitations around several models demonstrating relatively lower accuracies, indicating room for improvement. Furthermore, the reviewed papers often did not provide essential details such as the sizes of the models or the prediction times, which are crucial for evaluating the feasibility and efficiency of these approaches in real-world applications. More information is needed to ensure the ability to fully assess the practicality of deploying these models, especially in time-sensitive agricultural settings. Finally, there needs to be more comparative analysis across different methodologies. Most studies focus on a single approach rather than testing various methods to ascertain the most effective strategies for crop classification. Addressing these three areas, leveraging foundation models, enhancing interpretability, and conducting comparative analyses could lead to significant strides in applying machine learning in agriculture.

\section{Journal Contribution}
This research aims to try and overcome the limitations of previous studies by utilizing a comprehensive and multifaceted approach to crop classification through advanced machine learning and deep learning techniques while also providing explainability. The journal's contributions are as follows:
\begin{enumerate}

    \item \textbf{Diverse Data Acquisition: }The research uses a comprehensive dataset from multiple sources to acquire diverse images from the infield, such as drone imagery, phone images, and online images from various repositories. The dataset consists of various visual perspectives and contexts, crucial in developing a robust model capable of handling real-world variability in crop images.

    \item \textbf{Employment of Multiple Modelling Approaches: }The journal explored four distinct modelling approaches:
    \begin{itemize}
        \item Traditional machine learning models use features extracted using handcrafted feature extraction techniques like ORB, SIFT, and colour histograms combined with KNN and SVM classifiers.

        \item Deep learning models, specifically custom-designed CNNs and AlexNet Architecture, to leverage hierarchical feature extraction.

        \item Transfer learning using state-of-the-art architectures pre-trained on ImageNet, including EfficientNetV2, ResNet152V2, Xception, InceptionResNetV2, and MobileNetV3.

        \item Foundation models like YOLOv8 and DINOv2 to capitalize on their extensive pre-training and generalization capabilities.
    \end{itemize}

    \item \textbf{Multi-Dimensional Evaluation Criteria: }The model performance evaluation was comprehensive, focusing on recall, precision, accuracy, and confusion matrix analysis. This multi-dimensional evaluation thoroughly understood each model's strengths and weaknesses.

    \item \textbf{Focus on Explainability and Transparency: }A significant contribution of this study is its emphasis on Explainable AI (XAI) techniques, such as LIME, SHAP, and GradCAM. These tools provided insights into the decision-making processes of the models, enhancing their transparency and understandability. This is particularly important in agriculture, where decisions based on model predictions can have significant real-world impacts.

    \item \textbf{Practical Application and Relevance: }The journal's methodology is tailored to real-world applications in agriculture, emphasizing the need for accurate, reliable, and understandable crop classification models. By combining advanced machine learning techniques with a focus on model interpretability, the study contributes significantly to precision agriculture.
\end{enumerate}

\section{Methodology}
The methodology encompassed a comprehensive multi-stage process, including robust data acquisition, preparation, model development, evaluation, and explainability. Data was sourced from various means, including drone-captured imagery, phone videos and images, and online image repositories, to ensure a diverse and comprehensive dataset. The preparation phase involved video frame extraction, precise data cropping, annotation, and image combination to create a standardized dataset for subsequent modelling. Then, four modelling approaches were employed, beginning with feature extraction techniques such as ORB, SIFT and colour histogram analysis and then using the features to train traditional models for classification like KNN and SVM algorithms. Secondly, deep learning models, specifically custom-designed CNNs and AlexNet Architecture, were explored to exploit hierarchical feature extraction. Thirdly, transfer learning was utilized using architectures pre-trained on ImageNet, like EfficientNetV2, ResNet152V2, Xception, InceptionResNetV2, and MobileNetV3, to take advantage of pre-existing complex feature representations. Finally, Foundation models like YOLOv8 and DINOv2 were included to utilize their extensive pre-training and generalization capabilities. Moving next was the multi-dimensional Evaluation criteria, focusing on recall, precision, and accuracy metrics, alongside confusion matrix analysis for a comprehensive assessment of model performance. The interpretability and transparency of the models were assured through Explainable AI (XAI) techniques, with tools such as LIME, SHAP, and GradCAM providing insights into the decision-making processes, thus fostering a deeper understanding of model outcomes. The

\begin{figure}[h]
\centering
\includegraphics[width=0.8\textwidth]{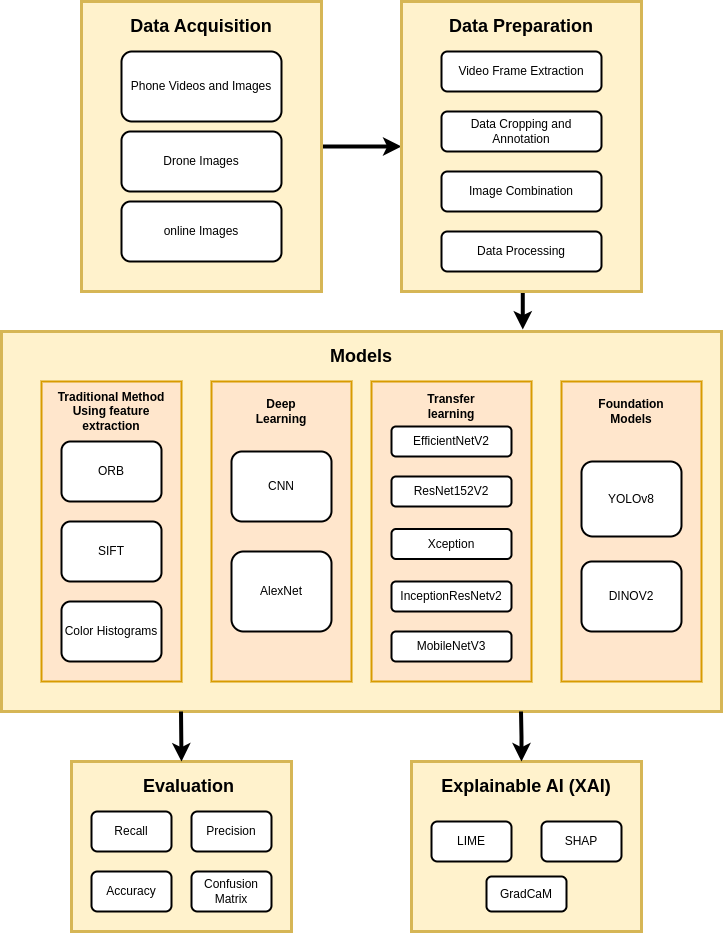}
\caption{Methodology flow Diagram}
\label{fig:cat}
\end{figure}

\subsection{Data Acquisition}
\begin{figure}[!htb]
\centering
\includegraphics[width=0.75\textwidth]{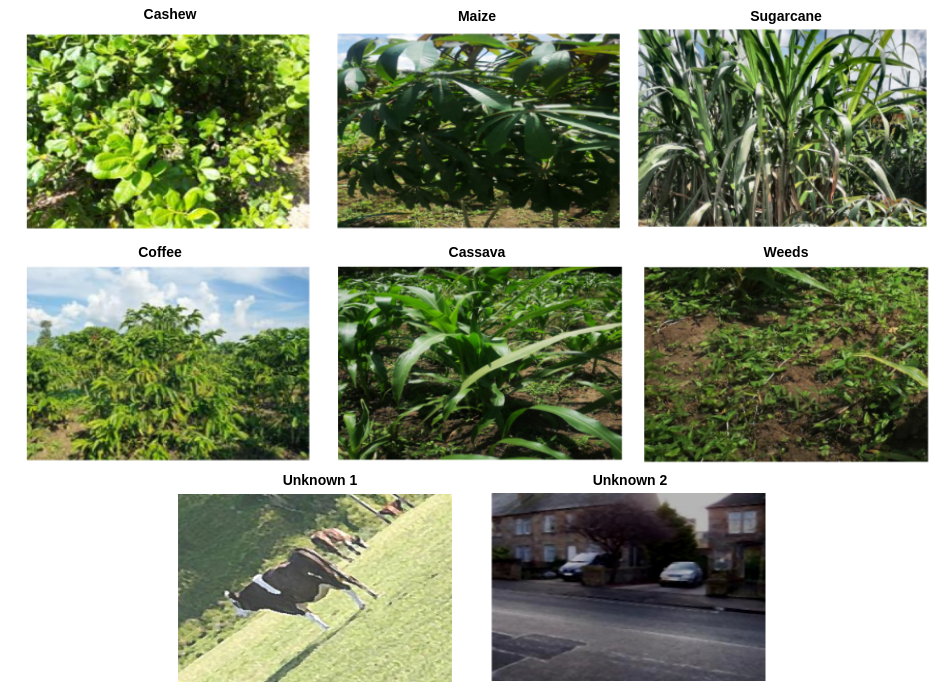}
\caption{Examples of images that represent all seven classes. For the unknown class, various images were used, including different types of images that do not belong to any of the six classes, namely cassava, sugarcane, maize, grass, cashew, and coffee.}
\label{fig:images}
\end{figure}

Seven classes were considered, including five crops (cassava, sugarcane, maize, grass, cashew, and coffee), weeds in the infield, and unknown images. The first phase of the methodology was focused on collecting a comprehensive dataset. For the drone-captured imagery, sample images from high-resolution aerial views and sides of cashew and coffee were picked from the large dataset collected in partnership with Makerere AI Lab, Uganda Marconi Lab, National Coffee Research Institute, and National Crops Resources Research Institute \cite{SANYA2024109952}. This data was published for the research community to study. In addition, phone videos were recorded in the infield garden, which contained three crops (cassava, sugarcane, and maize) and the weeds. The images extracted from the videos provided on-the-ground perspectives. Finally, the unknown images were extracted from online image repositories, contributing a breadth of pre-captured visual data. The collection process was designed to ensure a wide variety of visual angles, resolutions, and contexts to create a dataset. Table \ref{tab:image_data} shows all images acquired, which will be elaborated in the following subsection. Figure \ref{fig:images} shows all sample images for all seven classes.

\begin{table}[h]
\centering
\caption{Dataset Overview}
\label{tab:image_data}
\begin{tabular}{@{}lcccccc@{}}
\toprule
\makecell{Image Class} & \makecell{Total Images\\Acquired\\ \textbf{(A)}} & \makecell{Val Data\\Images\\Extracted\\from A \textbf{(B)}} & \makecell{Test Data\\Images\\Extracted\\from A \textbf{(C)}} & \makecell{Train Data\\Remaining\\data After\\A-B-C\textbf{(D)}} & \makecell{Actual Training\\Data After\\Augmentation\\of D \textbf{(E)}} & \makecell{Total Images\\After Data\\Augmentation\\B+C+E\textbf{(F)}} \\ 
\midrule
Cashew      & 330 & 32 & 50 & 248 & 500 & 582 \\
Coffee      & 308 & 32 & 50 & 226 & 500 & 582 \\
Cassava     & 291 & 32 & 50 & 209 & 500 & 582 \\
Maize       & 280 & 32 & 50 & 198 & 500 & 582 \\
Sugarcane   & 127 & 32 & 50 & 45  & 500 & 582 \\
Weeds       & 281 & 32 & 50 & 199 & 500 & 582 \\
Unknown     & 224 & 32 & 50 & 142 & 500 & 582 \\
\midrule
\makecell{\textbf{Total}} & \textbf{1841} & \textbf{224} & \textbf{350} & \textbf{1268} & \textbf{3500} & \textbf{4074} \\
\bottomrule
\end{tabular}
\end{table}

\subsection{Data Preparation}
Various data preparation techniques were employed after acquiring different corresponding data. The aim was to use well-prepared data. More emphasis was placed on creating good images from the collected infield videos. Several steps were taken to convert the provided video into a dataset of images, crop out specific categories of crops, add these images to the drone and online images, and address the slight imbalances from different classes.

The following detailed explanation outlines each step performed:

\begin{enumerate}
    \item \textbf{Video to Image Conversion: }The source video was converted into a dataset of images using the ffmpeg tool in Linux. The video generated 979 images by extracting four frames per second of Cassava, Maize, Sugarcane and Weeds, as shown in table \ref{tab:image_data} \textbf{(A)}.

    \item \textbf{Crop Extraction: }The GNU Image Manipulation Program (GIMP), an open-source tool available for Linux, was used to crop the images and focus on specific crop categories.

    \item \textbf{Categorization and Folder Grouping: }After extracting the crops from the images, they were sorted into categories. The images taken from videos belonged to four classes: Cassava, Maize, Sugarcane, and Weeds. Additionally, drone images of cashews and coffee were downloaded, and online images of known crops were included. In total, there were seven groups to be made. The images were then grouped into separate folders to organize the data, with each folder representing one of the seven classes (Cashew, Coffee, Maize, Cassava, Sugarcane, Weeds, and Unknown). This step aimed to establish distinct data for each category. The number of total images obtained is 1841, as shown in table \ref{tab:image_data} \textbf{(A)}.

    \item \textbf{Data Partitioning: }Since the dataset size was small and imbalanced, it was necessary to partition the data into three sets: training, testing, and validation. Additionally, removing the validation and test data before data manipulation or augmentation is a good practice. Each class was allocated 50 images for the test set, resulting in 350 images (50 images per class); this is shown in the table \ref{tab:image_data} \textbf{(C)}. Similarly, 32 images were allocated for each class in the validation set, resulting in 224 images (32 images per class); this is shown in the table \ref{tab:image_data} \textbf{(B)}. After allocating the test and validation sets, the remaining images were assigned to the training set. The number of images varied based on the initial class size and the number allocated for testing and validation. For the train set, 1268 original images were used after removing test and validation data as shown in the table \ref{tab:image_data} \textbf{(D)}.

    \item \textbf{Data Augmentation: }Data augmentation techniques were applied to address the small dataset size and further enhance the training set. The Albumentations Library, a powerful open-source library for image augmentation, was used for this purpose. Data augmentation involved applying various transformations to the existing training images, such as rotation, scaling, flipping, and other techniques. The number of total images obtained after augmenting the training set is 3500 images, as shown in the table \ref{tab:image_data} \textbf{(E)} The objective was to increase the number of images in the training set by oversampling each class to a minimum of 500.

\end{enumerate}

\begin{figure}[!htb]
\centering
\includegraphics[width=1\textwidth]{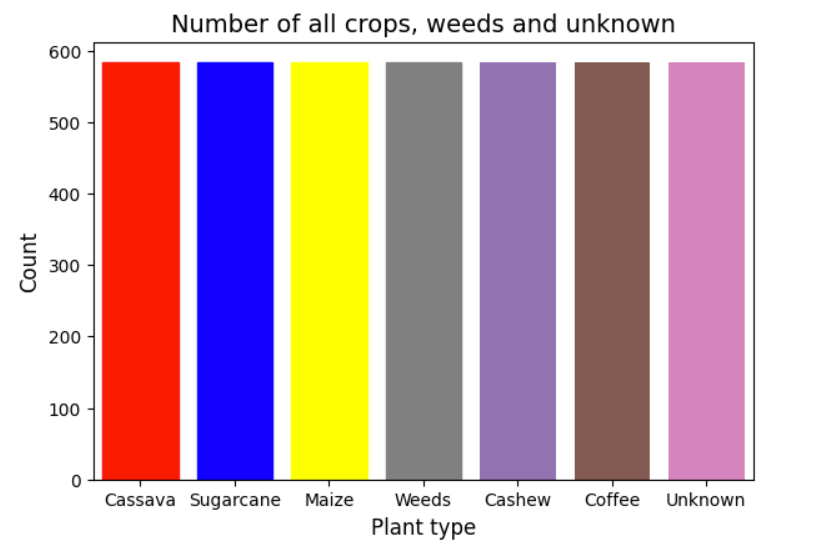}
\caption{Data samples from the developed dataset after augmentation cropping and annotating}
\label{fig:type1}
\end{figure}

Following the above procedure, the data set was prepared nicely, organized into separate folders for each crop category, and divided into training, testing, and validation sets. Data augmentation techniques were utilized to expand the training set size and address the imbalance in class distribution. The total number of images used is 4074 images, as shown in the table \ref{tab:image_data} \textbf{(D)}, and this is also represented in figure \ref{fig:type1}. The dataset has been published on Harvard Dataverse for research purposes. However, due to its size, it was divided into two parts. The first part consists of five classes in the training set which include cashew, cassava, coffee, maize, and sugarcane. This data is available  \href{https://doi.org/10.7910/DVN/J0OS9R}{here on this DOI}. The second part includes the remaining classes in the training set (weeds and unknown), validation, and test data. This data is available \href{https://doi.org/10.7910/DVN/VQOO17}{ here on this DOI}. To use the dataset, both parts need to be downloaded and combined with having preprocessed data ready for modelling.

\subsection{Classification Models}
After preparing the data, the next task was to develop a crop classifier using the data. Four different modelling approaches were considered based on a literature review to ensure that the best model is selected. The first method involved developing a classifier using important features or information extracted using popular algorithms such as SIFT and ORB. An alternative feature extraction technique, image colour histograms, was also explored. The extracted features were then classified using K-nearest neighbour (KNN) and SVM. Both SVM and KNN were used for comparison reasons. The second approach involved using a Deep Neural Network (DNN). For this, a custom Convolution Neural Network (CNN) for classification was developed, and a famous CNN architecture called AlexNet was used. Both the custom CNN and AlexNet architecture were used for comparison for the Deep Learning approach. The third approach involved using transfer learning on pre-trained models using some of the best models trained using Imagenet data, including EfficientNetV2, ResNet152V2, Xception, InceptionResNetv2, and MobileNetV3. The last approach was to use foundation models. These models are trained on massive datasets, and two models used are: YOLOv8 and DINOv2.

\subsubsection{\textbf{Traditional Method Using Feature Extraction:}}

The objective was to explore different techniques for extracting meaningful information from images using feature extraction. Three standout techniques were SIFT, ORB, and colour histogram, which were then used for image classification. KNN and SVM were chosen as the classical machine learning algorithms for their simplicity and effectiveness in image classification tasks. For each image in the dataset, edge detection was performed, and the SIFT and ORB algorithms were applied to extract feature descriptors that capture distinctive characteristics of the images. Feature descriptors were collected and organized into a data frame, each row representing an image. Each image was represented as a row in a data frame for the colour histogram. The features in the data frame were the colour histograms with a fixed number of bins. This method enabled a concise representation of colour distribution for each image. For KNN, the hyperparameter tuning process was performed to determine the optimal hyperparameter k value in the KNN algorithm. The value of k was systematically varied, and the model's performance was evaluated on the validation set. The iterative process was employed to search for the best k value that yielded the most accurate classification results.

To evaluate the effectiveness of the KNN and SVM algorithms, a dataset represented in Table \ref{tab:image_data} was used, considering only the training and testing sets. Using it like that allowed the model to learn from a significant portion of the data while still being able to evaluate its performance on unseen samples. Additionally, feature normalization was performed before training the KNN and SVM models to ensure that all features had similar scales and distributions. This step was necessary to prevent any bias towards features with larger magnitudes and ultimately improve the accuracy of the classifier. The results obtained from different feature extraction methods were compared to identify the most effective approach for classifying images using the KNN and SVM algorithms.

\begin{figure}[!htb]
\centering
\includegraphics[width=1\textwidth]{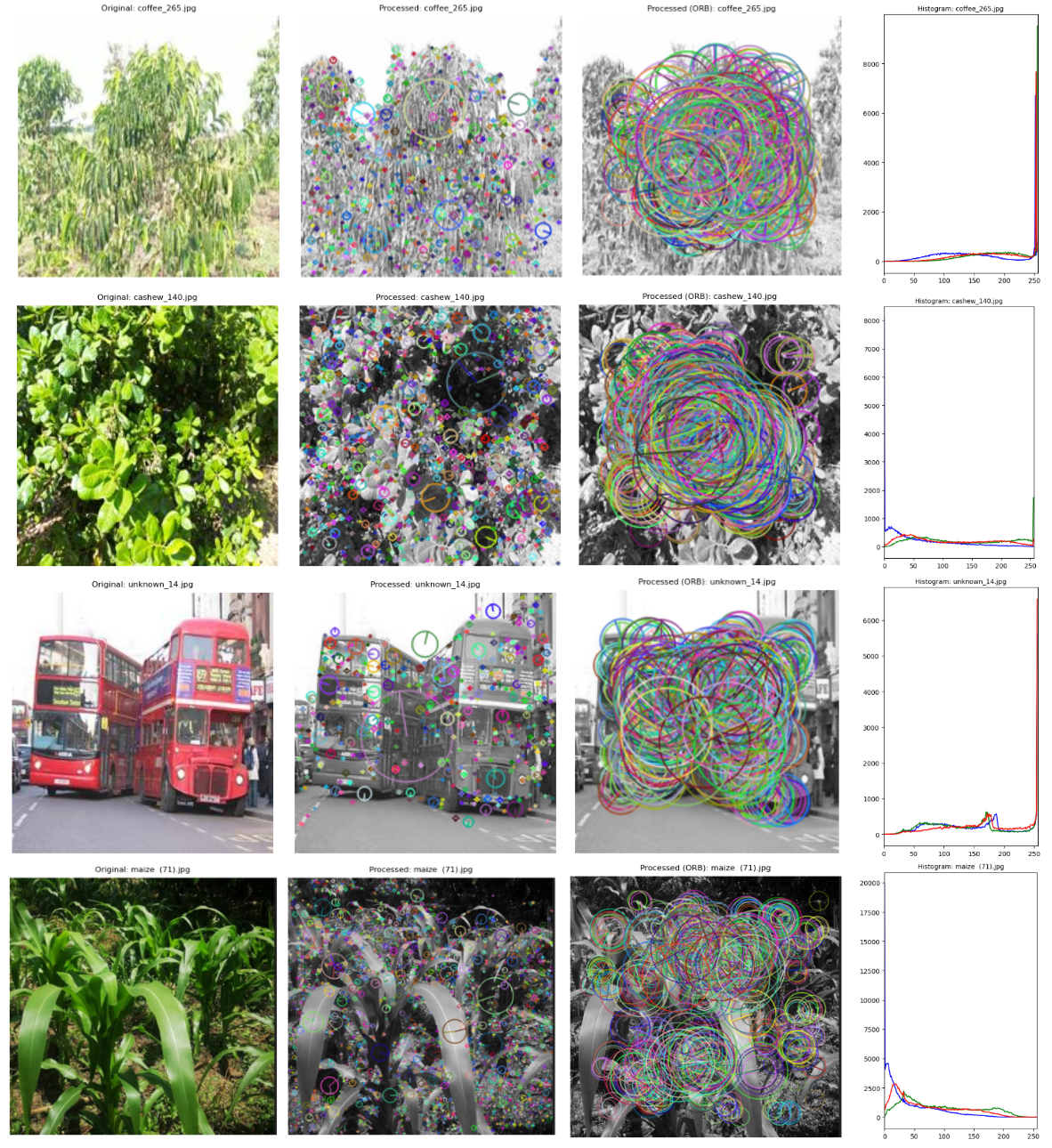}
\caption{Sample features representation using SIFT (second column), ORB (Third Column), and Color Histogram (Fourth Column) that can be extracted from images in the first column.}
\label{fig:type5}
\end{figure}

\begin{enumerate}
    \item \textbf{SIFT Algorithm: }Image feature points are valuable for image identification and classification, as they possess unique characteristics that can distinguish one image from another. The SIFT algorithm identifies key points across different scale-spaces and calculates their directions. These key points correspond to distinctive features such as corners and mutations, enabling effective image classification based on their characteristics. This approach demonstrates robustness against various image distortions, including rotation, scaling, and noise \cite{wang2021improved}. However, the accuracy of image classification relying on feature points is influenced by the quality of the extracted feature points. In this study, the SIFT algorithm is employed for feature extraction due to its ability to generate high-quality feature points, resulting in improved classification accuracy compared to other algorithms like SURF \cite{12}. Examples of these features are shown in image \ref{fig:type5} second column. 

    \item \textbf{ORB Algorithm: }In addition to SIFT, the Oriented FAST and Rotated BRIEF (ORB) algorithm is another popular algorithm for extracting key points and descriptors. ORB is known for its efficiency and robustness in various computer vision applications, including image classification. Like SIFT, ORB identifies local features or keypoints in an image invariant to scale and rotation\cite{imsaengsuk2021feature}. These keypoints are characterized by quantitative descriptors, which play a vital role in image classification tasks\cite{imsaengsuk2021feature}. These descriptors are designed to be invariant to different transformations that may alter the appearance of the same object. The scale space is a fundamental concept in ORB, used to represent the image. It is obtained by convolving the Gaussian function with the original image. The Difference of Gaussian (DoG) is then computed by subtracting the convolved images obtained at two different scales\cite{dai2023improved}. This process is performed at multiple octaves in the Gaussian Pyramid, allowing the algorithm to capture features at different scales. Once the DoG is calculated, the algorithm searches for local extrema over scale and space. This involves comparing a pixel with its neighbours in the current scale and neighbouring pixels in adjacent scales. A pixel identified as a local extrema is considered a potential keypoint representing a distinctive feature in the image \cite{pham2017color}. The ORB algorithm provides an efficient and effective approach for extracting key points and descriptors from images. By utilizing scale invariance and local extrema, ORB enables robust feature extraction that is valuable for image classification tasks \cite{shamla2023feature}. Examples of these features are shown in image \ref{fig:type5} third column.

    \item \textbf{Color Histogram: } In addition to key point extraction techniques such as SIFT and ORB, colour histogram-based methods have been widely used in image classification tasks. A colour histogram represents the distribution of colours in an image by quantizing the colour space into bins and counting the occurrences of colours within each bin \cite{sahann2021histogram}. Constructing a colour histogram involves dividing the image into small regions or pixels and extracting the colour information from each region. The colour space, such as RGB (Red, Green, Blue) or HSV (Hue, Saturation, Value), can represent the colours \cite{yoo2022extraction}. Each pixel's colour values are mapped to the corresponding colour space, and the frequencies of colour occurrences are calculated for each bin. The colour histogram provides a concise representation of the image's colour distribution, capturing important colour characteristics that can be used for classification. It allows recognition of images based on their colour properties, such as dominant colours, colour contrasts, or colour patterns. This approach is beneficial when the colour information significantly discriminates between different classes or categories of images \cite{sergyan2008color}. Various algorithms and methods can be employed to utilize colour histograms for image classification. Commonly used techniques include histogram intersection, chi-square distance, or Euclidean distance, which measure the similarity between the colour histograms of different images \cite{chugh2022image}. Machine learning algorithms such as Support Vector Machines (SVM) or Random Forests can be trained on these histogram-based features to classify images into different categories \cite{sada2018histogram}. In summary, colour histograms provide a valuable representation of an image's colour distribution and have been successfully utilized in image classification tasks. By capturing essential colour information, they offer a practical approach to discriminate between different classes based on their colour characteristics. Examples of these features are shown in image \ref{fig:type5} fourth column.
\end{enumerate}

\subsubsection{\textbf{Classification using a CNN: }}
CNNs have significantly advanced image classification tasks, showing remarkable efficacy across various fields, including agriculture. Mimicking the human visual cortex, CNNs autonomously learn and extract hierarchical image features. Their architecture comprises convolutional layers to detect local patterns, pooling layers to reduce spatial dimensions, and fully connected layers to represent and classify high-level features. This structure effectively captures complex spatial dependencies in images, which is particularly useful for crop classification tasks.

\begin{figure}[h]
\centering
\includegraphics[width=1\textwidth]{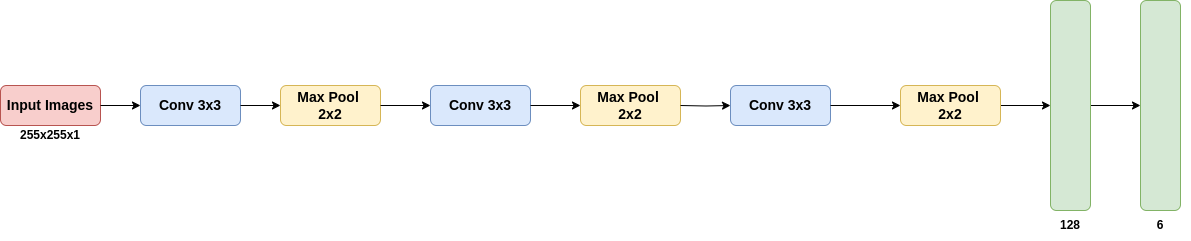}
\caption{Custom Designed CNN model for classification}
\label{fig:CNN}
\end{figure}

This study explicitly developed a custom CNN architecture for crop differentiation. The CNN comprises multiple stacked layers, including convolutional layers with adjustable filters for feature extraction, pooling layers for downsampling, and fully connected layers. Hyperparameters such as the number of filters and neurons were adjusted experimentally to optimize performance. The Rectified Linear Unit (ReLU) was used as the activation function to introduce non-linearity into the network. Additionally, the model's performance was evaluated by training it on our large enough dataset and testing it on validation data. This iterative process aimed to improve the accuracy, precision, and other metrics relevant to the task by modifying the architecture and hyperparameters of the model. The goal was to fine-tune the CNN to its optimal configuration, maximizing its efficacy in the specified application. Figure \ref{fig:CNN} shows the visual representation of the custom-designed CNN model.

In addition to the custom CNN architecture, this study also explored the application of the AlexNet architecture for crop classification. AlexNet is a well-known deep-learning architecture famous for recognizing complex patterns in images. Its structure includes five convolutional layers and three fully connected layers, which make it proficient at identifying intricate image features. Critical aspects of AlexNet, like its use of ReLU activation functions and overlapping pooling, significantly reduce the error rate in image classification tasks. For the crop classifier, AlexNet was chosen due to its ability to handle complex image datasets. The training involved feeding a substantial number of data to the network, allowing it to learn distinctive features of different crop types. The architecture's deep layers and large learning capacity can make differentiating subtle differences in crop images easier. 

\subsubsection{\textbf{Transfer Learning: }}

Transfer learning is actually a technique within the field of deep learning, not an alternative to it. It involves taking a pre-trained deep learning model and adapting it to a new but related problem. These pre-trained models are often used in deep learning for computer vision and natural language processing tasks. Developing custom neural network models for these tasks requires vast computing and time resources. Pre-trained models can improve related problems significantly and provide huge jumps in skill. This approach has gained significant popularity due to its benefits, especially when compared to training a deep learning model from scratch.  Machine learning practitioners prefer to use pre-trained models for such tasks.

One of the most significant advantages of transfer learning is that it does not require the large datasets necessary for training a deep learning model from scratch. The pre-trained model has already learned a lot of useful features from its initial training, which can be applied to the new task, even with a relatively small dataset. Secondly, Since the model is already trained on a large dataset, adapting it to a new task requires much less computational time and less time for fine-tuning the model to find the perfect parameters. This is particularly beneficial when resources or time are limited. Transfer learning can perform better tasks where the available data is limited. The pre-trained model brings in knowledge from a related task, which can be especially helpful when the new task needs more data to train a model effectively from scratch. When working with small datasets, there is a high risk of overfitting if a model is trained from scratch. Transfer learning also helps mitigate this risk since the model has already learned generalizable features. Pre-trained models, especially those trained on large and diverse datasets like ImageNet, have learned complex feature representations. Additionally, it allows these advanced features to be leveraged in tasks that may need to be developed independently due to data or resource constraints. Transfer learning is instrumental in domain adaptation, where the task is to apply knowledge learned in one domain to a different but related domain.

For crop classification, transfer learning capitalizes on the learned features often transferable to its visual task. By leveraging models pre-trained on the ImageNet dataset, we capitalize on the extensive feature extraction capabilities already inherent in these networks. Using pre-trained models is advantageous for the task as they have already acquired a rich set of image features, including basic shapes, edges, and complex patterns. The models employed for this purpose are some of the best in classification, including EfficientNetV2, ResNet152V2, Xception, InceptionResNetV2, and MobileNetV3, which are advanced neural network architectures that have achieved state-of-the-art results on benchmark datasets.

\begin{itemize}
    \item EfficientNetV2 is known for its scalability and efficiency. It builds upon the original EfficientNet design principles but includes additional optimizations that allow it to train faster and perform better. Its efficiency can be critical for crop classification in managing the computational costs while maintaining high accuracy.

    \item ResNet152V2 is a variant of the original ResNet architecture, which includes 152 layers. It utilizes residual connections to train such deep networks by enabling the flow of gradients directly through the layers. This can help learn complex patterns in agricultural data without the vanishing gradient problem.

    \item Xception stands for "Extreme Inception," which modifies the Inception architecture by replacing Inception modules with depthwise separable convolutions. This could improve the performance of crop images by focusing on cross-channel correlations and capturing spatial hierarchies more effectively.

    \item InceptionResNetV2 combines Inception architectures' benefits with ResNet's residual connections. This hybrid approach can provide the model with a richer representation of features, which helps distinguish between different types of crops that may have subtle visual differences.

    \item MobileNetV3 is designed to be lightweight and efficient, making it suitable for mobile and edge devices. It uses techniques from previous MobileNet models, including depthwise separable convolutions, to maintain high accuracy with a lower computational cost. For real-time crop classification on handheld devices, MobileNetV3 can be an excellent choice.
\end{itemize}

By leveraging these pre-trained models through transfer learning, the crop classification system can achieve high levels of accuracy without the need to train a model from scratch. The vast amount of feature representations learned from ImageNet enables these models to understand and classify complex patterns in crop imagery, even when the dataset is relatively small. This is particularly advantageous in precision agriculture, where it is essential to correctly identify and classify crop types for monitoring crop health, yield estimation, and resource management.

\subsubsection{\textbf{Foundation Models: }}
Foundation models are deep learning models pre-trained on a large amount of data. They can be fine-tuned for specific tasks and are also great starting points for many applications. They can learn general representations that can be applied to many downstream tasks. The term "foundation" reflects their broad applicability and the fact that they are the basis for specialized models. The main difference between foundation models and transfer learning is the scale and capability of the models. Foundation models are usually cutting-edge deep learning models that provide state-of-the-art performance and require significant computational resources to train and fine-tune. On the other hand, transfer learning can be achieved with less computational effort and may not necessarily require models as advanced as foundation models.

Two foundation models were trained for this task YOLOv8 (You Only Look Once, version 8) and DINOv2 (Knowledge Distillation with No Labels, version 2). YOLOv8 is an evolution of the YOLO family, known for its real-time object detection capabilities, which can be crucial for identifying and classifying crops in images quickly and accurately. Its ability to process images in a single pass enables rapid detection of multiple crop types within a scene, making it suitable for monitoring crop health or detecting weed species in real-time. On the other hand, DINOv2, which also stands for Detection with Transformers, represents a newer paradigm in deep learning that leverages the Transformer architecture, commonly seen in natural language processing, for computer vision tasks. This model is adept at handling large-scale and complex image datasets, making it suitable for distinguishing between subtle variations in crop species and detecting patterns indicative of diseases or stress conditions in crops. When fine-tuning these foundation models for crop classification, the large and diverse pre-training on ImageNet provides them with a substantial visual vocabulary, which can be refined to the specifics of agricultural imagery. By adjusting the final layers of these models to focus on the relevant crop types and training them on labelled crop data, these foundation models can be highly effective for precision agriculture tasks, providing detailed insights into crop variability and enhancing yield predictions.

\subsubsection{\textbf{Evaluation Metrics:}}
To assess the performance of the crop classification models, we employed a suite of standard evaluation metrics, each providing insights into different aspects of the model's predictive capabilities. Their summary and equations follow where $TP$ is the true positive, $FN$ is the false negative,  and $FP$ is the false positive.


\begin{itemize}
    \item Recall measures the ability of the model to identify all relevant instances of a particular class. In crop classification, a high recall rate indicates that the model can identify most of the crop instances from the data, which is crucial for applications where missing a particular crop type could have significant consequences, such as pest detection or yield estimation.

    \begin{equation}\text{recall} = \frac{TP}{TP+FN}\end{equation}

    \item Precision quantifies the model's accuracy in labelling an instance belonging to a particular crop class. A high precision score reflects the model's ability to minimize false positives, ensuring it is classified correctly when a crop type is identified. This is especially important in precision agriculture, where misclassification can lead to inappropriate treatment or harvesting strategies.

    \begin{equation}\text{precision} = \frac{TP}{TP+FP}\end{equation}

    \item Accuracy is the most intuitive performance measure, and it is simply a ratio of correctly predicted observations to the total observations. It gives a general sense of how often the model is correct across all classes, but it may not be as informative in imbalanced datasets where some crop types are much more common than others.

    \begin{equation}
    \text{Accuracy} = \frac{\text{Number of Correct Predictions}}{\text{Total Number of Predictions}}
    \end{equation}

    \item Confusion Matrix offers a detailed model performance breakdown, showing the number of correct and incorrect predictions made for each class. This matrix is particularly useful for identifying specific classes where the model may be underperforming and need additional training data or re-tuning of the model parameters.
\end{itemize}

These metrics were computed using a validation dataset the model did not see during training. By analyzing these metrics, we can gauge the robustness and reliability of our crop classification models, ensuring they are both accurate and practical for deployment in agricultural settings.

\subsubsection{\textbf{Explainable AI (XAI): }}
In advanced machine learning and AI, achieving high-performance metrics across various models, including traditional machine learning, deep learning architectures, transfer learning, and foundation models, marks a significant accomplishment. However, these models' complexity and often opaque nature necessitate an understanding beyond mere performance metrics. This is where Explainable Artificial Intelligence (XAI) becomes a crucial component, enhancing the transparency and interpretability of these sophisticated models \cite{murindanyi2023interpretable}.

Applying XAI is not just a matter of academic interest; it is a practical necessity in fields where AI-driven decisions, such as medicine and agriculture, can have significant impacts. Despite their effectiveness, advanced models are frequently critiqued for their "black-box" nature, where the reasoning behind their predictions is not readily apparent. This lack of transparency can be a major hurdle in deploying AI solutions in real-world scenarios where trust, understanding, and ethical considerations are paramount.

In an effort to build trust and understanding of our crop classification models, we incorporated Explainable AI (XAI) techniques. XAI aims to make the operation of complex models transparent and comprehensible to humans. For this purpose, we utilized three prominent methods: LIME, SHAP, and Grad-CAM.

\begin{itemize}
    \item \textbf{LIME (Local Interpretable Model-agnostic Explanations)} provides insights by perturbing the input data and observing the corresponding prediction variation. This technique helps understand the model's behaviour on an individual prediction basis, offering a local perspective of the model's decision-making process \cite{murindanyi2023explainable}. In crop classification, LIME can indicate which pixels and features of the image influence the model's prediction for a particular crop type.

    \item \textbf{SHAP (SHapley Additive exPlanations)}, grounded in game theory, assigns each feature an importance value for a particular prediction. This method ensures a consistent and fair contribution allocation across all features, allowing us to identify which characteristics of the crop images most significantly impact the model's output. SHAP values can be particularly enlightening when determining the factors differentiating one crop.

    \item \textbf{Grad-CAM (Gradient-weighted Class Activation Mapping)} utilizes the gradients of any target concept, flowing into the final convolutional layer to produce a coarse localization map highlighting the important regions in the image for predicting the concept. In our context, Grad-CAM visually demonstrates where the model is focusing its attention within the image during the classification process. This can be crucial for validating whether the model is considering the appropriate parts of the image, such as the leaves or fruits of crops, rather than irrelevant background features.
\end{itemize}

By applying these XAI techniques, we not only bolster the interpretability of our models but also ensure they are making decisions based on relevant features. This is essential for real-world applications where the reasoning behind a model's decision must be justifiable and understandable to domain experts.

\section{Experimental Results and Discussion}
In this section, the results and analysis of the work on the models and their performance are presented. 

\subsection{First Approach: Feature Extraction with Traditional Algorithms and Classical ML Models}
For the traditional feature extraction method, after performing feature selection and preprocessing, the features were prepared for training by partitioning them into a 70\% training set and a 30\% test set. Both K-nearest neighbours and SVM algorithms were trained using the prepared features. For KNN, a prediction process that considered five neighbours was utilized. The training set comprised 954 images, while the test set contained 410 images.

\begin{table}[ht]
\centering
\caption{Performance metrics of different feature extractors with KNN and SVM models}
\label{tab:1}
\resizebox{\textwidth}{!}{%
\begin{tabular}{@{}lllllllllll@{}}
\toprule
\multicolumn{1}{c}{\textbf{Feature Extractors}} & \multicolumn{1}{c}{\textbf{Models}} & \multicolumn{1}{c}{\textbf{Tr. Acc}} & \multicolumn{1}{c}{\textbf{Test Acc}} & \multicolumn{1}{c}{\textbf{Prediction}} & \multicolumn{1}{c}{\textbf{Recall}} & \multicolumn{1}{c}{\textbf{F1-Score}} & \multicolumn{1}{c}{\textbf{Size/MB}} & \multicolumn{1}{c}{\textbf{Time/seconds}} \\ 
\midrule
\multicolumn{1}{l}{SIFT} & KNN & 91\% & 85\% & 85\% & 85\% & 85\% & 0.286 & 0.101164 \\
\multicolumn{1}{l}{} & SVM & 85\% & 85\% & 85\% & 85\% & 85\% & 0.132 & 0.049114\\
\multicolumn{1}{l}{ORB} & KNN & 66\% & 46\% & 48\% & 46\% & 47\% & 0.286 & 0.109706\\
\multicolumn{1}{l}{} & SVM & 48\% & 46\% & 45\% & 46\% & 46\% & 0.296 & 0.106158\\
\multicolumn{1}{l}{Color Histogram} & KNN & 95\% & 90\% & 91\% & 91\% & 90\% & 5.60 & 0.0162\\
\multicolumn{1}{l}{} & SVM & 96\% & 93\% & 94\% & 94\% & 94\% & 5.88 & 0.0027\\ \bottomrule
\end{tabular}%
}
\end{table}

The findings, as summarized in Table \ref{tab:1}, indicate notable differences in performance among various feature extractors and machine learning models. Firstly, combining Color Histogram with SVM achieved the highest accuracy in training at 96\% and testing at 93\%. It also showed impressive precision, recall, and F1 scores above 90\%. Additionally, this combination exhibited a very short processing time of 0.0027 seconds but required a larger model size of 5.88 MB. These results show that combining Color Histogram with SVM is highly effective for classifying this specific crop, providing an excellent balance of accuracy and computational efficiency.

Secondly, The ORB feature extractor exhibited the lowest performance results, particularly when combined with the SVM model. As a result, its test accuracy was only 46\%. Although the ORB extractor has a small model size and processing time, making it computationally efficient, it did not accurately classify crops. This implies that ORB might not be suitable for complex image classification tasks, such as crop classification, where the features could be subtle and nuanced.

Finally, the performance of SIFT was more balanced, achieving a consistent 85\% across accuracy, precision, recall, and F1-score metrics with both KNN and SVM models. However, despite moderate processing time and model size, the Color Histogram demonstrated higher performance than SIFT, suggesting room for improvement.

Choosing the right combination of feature extraction method and machine learning model is crucial for crop classification applications using this method. The Color Histogram with SVM's superior performance highlights its potential in practical agricultural technology applications where accurate crop classification is crucial. However, the choice of feature extractor and model should also consider the computational resources available and the specific requirements of the task at hand.

In conclusion, this approach involves using predefined handcrafted features designed to capture specific aspects of the data in our aspect different crop data. However, it has limitations because it requires domain expertise to know how to use those feature extractors and often fails to capture high-level and complex patterns in data, particularly in varied and dynamic environments. On the other hand, deep learning automatically learns feature representations from data, making it more adaptable and efficient in capturing complex and abstract patterns. Furthermore, deep learning models are generally more robust to variations in input data, such as changes in lighting, orientation, and scale. This is because they can learn invariant features not easily captured by traditional handcrafted feature extraction methods.

\subsection{Second Approach: DNN}
Deep learning enables direct input of raw data into the model, resulting in end-to-end learning. This contrasts traditional methods requiring separate stages, including feature extraction, selection, and classification. Each stage requires fine-tuning and optimization, and the process is often a time-consuming and expertise-intensive process. Deep learning eliminates all these and automatically allows for faster development and deployment of models.

\begin{table}[ht]
\centering
\caption{Performance metrics of different models}
\label{tab:2}
\resizebox{\textwidth}{!}{%
\begin{tabular}{@{}lllllllllllll@{}}
\toprule
\multicolumn{1}{c}{\textbf{Models}} & \multicolumn{1}{c}{\textbf{Tr. Acc}} & \multicolumn{1}{c}{\textbf{Val. Acc}} & \multicolumn{1}{c}{\textbf{Train Loss}} & \multicolumn{1}{c}{\textbf{Val Loss}} & \multicolumn{1}{c}{\textbf{Test Acc}} & \multicolumn{1}{c}{\textbf{Precision}} & \multicolumn{1}{c}{\textbf{Recall}} & \multicolumn{1}{c}{\textbf{F1-Score}} & \multicolumn{1}{c}{\textbf{Size/ MB}} & \multicolumn{1}{c}{\textbf{Time/ second}}\\ 
\midrule
\multicolumn{1}{l}{CNN} & 96\% & 94\% & 0.08 & 0.40 & 92\% & 92\% & 92\% & 92\% & 85.08  & 0.0755 \\
\multicolumn{1}{l}{AlexNet} & 76\% & 66\% & 0.77 & 1.18 & 65\% & 76\% & 65\% & 65\% & 714.11 & 0.0568\\
\bottomrule
\end{tabular}%
}
\end{table}

The experiments on deep learning models for crop classification provide a convincing comparison between a custom CNN and the well-known deep learning neural network AlexNet architecture. The findings, presented in Table \ref{tab:2}, demonstrate the advantages and drawbacks of these methods in managing intricate image classification tasks, such as identifying different crops.

The custom CNN model displayed outstanding performance by surpassing AlexNet in almost all metrics. The model achieved a high training accuracy of 96\% and a validation accuracy of 94\% while maintaining a low training loss of 0.08 and a moderate validation loss of 0.40. Notably, the model's test accuracy of 92\% highlights its capability to perform well on unknown data. The CNN model also consistently performed with precision, recall, and F1-score metrics at 92\%. It has a substantial size of 85.08 MB and a prediction time of 0.0755 seconds. These results suggest that the custom CNN is highly effective for this specific application, balancing accuracy and computational efficiency.

On the other hand, AlexNet did not perform well in this particular task. Its training and validation accuracies were substantially lower, at 76\% and 66\%, respectively. The model also had a higher training loss of 0.77, validation losses of 1.18, and a lower test accuracy of 65\%. The precision of the model was 76\%, which is relatively high compared to its recall and F1-score at 65\%. However, compared to the result of the custom CNN model, these scores were lower, indicating some challenges with model generalization and consistency. Although the prediction time of the AlexNet model was slightly less at 0.0568 seconds, its size was considerably larger at 714.11 MB.

In conclusion, the results indicate that the custom CNN architecture developed for crop classification is highly effective. While the AlexNet model is a powerful and widely used architecture in deep learning, it was less effective than a custom-designed CNN for this specific task. This highlights the importance of selecting and customizing models to achieve high performance in specialized areas. The custom CNN's ability to maintain high accuracy while keeping computational demands reasonable makes it a promising tool for practical applications in agriculture and related fields, where efficient and accurate classification of crop types is a critical factor.

\subsection{Comparison Between the first approach and the second approach}
The first approach used for this project was to use traditional feature extractors such as SIFT, ORB, and Color Histogram, combined with classical machine learning models like KNN and SVM models to classify various crop types in image classification tasks. This approach emphasizes the significance of feature selection, particularly the quality and relevance of manually chosen features, such as colour, to ensure effective classification. The performance of different feature extractors and classifier combinations varied significantly, with Color Histogram and SVM showing the highest effectiveness. This approach requires less computational power than deep learning models, as indicated by their smaller model sizes and faster processing times. However, it may need more depth of learning and generalization capabilities that deep learning offers, particularly in cases where crop images have complex and subtle features.

A custom CNN and the AlexNet architecture were utilized for the deep learning approach, demonstrating the power of convolutional neural networks in automatically learning hierarchical feature representations from the images. The custom CNN, developed specifically for this task, outperformed AlexNet, showing higher accuracy, precision, recall, and F1 scores. Additionally, DNNs, especially CNNs, excel in handling high-dimensional data and capturing intricate image patterns, which is critical for accurate crop classification. Although using deep learning models for crop classification has shown promising results, it comes with a trade-off of higher computational requirements. This is evident in the larger model sizes and the need for more processing power. While AlexNet is a well-known architecture, it did not perform optimally in this specific application. Thus, customized models may better suit specialized tasks like crop classification.

\subsection{Third Approach: Transfer learning}
Transfer learning as a technique that uses a pre-trained model that has already learned important features from a related task. It significantly reduces the computational time and effort needed to fine-tune the model for a new task and mitigates the risk of overfitting when working with small datasets. Since the pre-trained model has already learned generalizable features, it produces better results.

\begin{table}[ht]
\centering
\caption{Performance metrics of different transfer learning models}
\label{tab:3}
\resizebox{\textwidth}{!}{%
\begin{tabular}{@{}lllllllllllll@{}}
\toprule
\multicolumn{1}{c}{\textbf{Models}} & \multicolumn{1}{c}{\textbf{Tr. Acc}} & \multicolumn{1}{c}{\textbf{Val. Acc}} & \multicolumn{1}{c}{\textbf{Train Loss}} & \multicolumn{1}{c}{\textbf{Val Loss}} & \multicolumn{1}{c}{\textbf{Test acc}} & \multicolumn{1}{c}{\textbf{Precision}} & \multicolumn{1}{c}{\textbf{Recall}} & \multicolumn{1}{c}{\textbf{F1-score}} & \multicolumn{1}{c}{\textbf{size/ MB}} & \multicolumn{1}{c}{\textbf{time/ second}}\\ 
\midrule
\multicolumn{1}{l}{EfficientNetV2} & 45\% & 54\% & 1.63 & 1.53 & 57\% & 73\% & 57\% & 57\% & 451.01  & 0.1136 \\
\multicolumn{1}{l}{ResNet152V2} & 99\% & 97\% & 0.01 & 0.06 & 98\% & 98\% & 98\% & 98\% & 223.68 & 0.0914\\
\multicolumn{1}{l}{Xception} & 99\% & 99\% & 0.01 & 0.04 & 98\% & 98\% & 98\% & 98\% & 80.03 & 0.0633\\
\multicolumn{1}{l}{InceptionResNetV2} & 100\% & 96\% & 0.00084 & 0.09 & 93\% & 95\% & 93\% & 93\% & 227.42 & 4.103\\
\multicolumn{1}{l}{MobileNetV3} & 53\% & 60\% & 1.42 & 1.34 & 60\% & 68\% & 60\% & 60\% & 11.90 & 1.2746\\
\bottomrule
\end{tabular}%
}
\end{table}

As your results show, implementing transfer learning models for crop classification introduces a third approach that leverages pre-trained neural networks. This method involves using these models for the task of crop classification. The performance of various transfer learning models, as detailed in Table \ref{tab:3}, showcases the effectiveness and diversity of this approach. Crucially, all models were trained for 30 epochs to have a consistent number for comparison purposes.

EfficientNetV2, with a test accuracy of 57\% and a precision of 73\%, did not perform as well as expected for crop classification, though its full capability might be unlocked with more extensive training. This model, engineered for a balance between accuracy and efficiency, underperformed due to limited training. Only 30 epochs were run, which were insufficient for optimal convergence. In contrast, MobileNetV3, known for its small size and low computational demands, achieved a moderate accuracy of 60\% and precision of 68\%. This architecture is designed to be efficient in mobile and resource-constrained settings. However, it showed signs of slow convergence and required at least 300 epochs to reach a minimum validation accuracy of 80\%. Despite its compactness, this model's extensive training requirements suggest it may not be suitable for crop classification tasks. The moderate performance of both models suggests that there may be a compromise between their lightweight design and the depth of feature analysis required for this task. Therefore, improving the duration and number of epochs of their training could enhance their ability to distinguish the intricate details of crop images.

In contrast, ResNet152V2, Xception, and InceptionResNetV2 demonstrated remarkable performance. ResNet152V2 and Xception both achieved very high training and validation accuracies of 99\% and 97\% for ResNet152V2, 99\% for Xception and test accuracies of 98\%. In contrast, ResNet152V2, Xception, and InceptionResNetV2 demonstrated outstanding performance. ResNet152V2 and Xception achieved a very high training accuracy of 99\%, and validation accuracies of 97\% for ResNet152V2, 99\% for Xception, and test accuracies of 98\% for both. These models also showed excellent precision, recall, and F1 scores, all hovering around 98\%. The InceptionResNetV2 model achieved a perfect training accuracy of 100\%, a high validation accuracy of 96\% and a test accuracy of 93\%. Additionally, the model exhibited high precision, recall, and F1 scores. These impressive results demonstrate the effectiveness of deep neural networks in accurately classifying crops despite the larger size and complexity of the model. It highlights how these networks can learn intricate and subtle features required for precise classification.

Based on all the models for crop classification, the Xception can be identified as the top-performing model. This remarkable model has achieved exceptional results during the training and validation stages, with accuracies of 99\%. Additionally, it has demonstrated a very high level of accuracy of 98\% when tested on unseen data. These metrics indicate that the Xception model has effectively learned from the training data and has been able to generalize well to new, previously unseen data. Moreover, its precision, recall, and F1-score are 98\%, demonstrating its robustness in accurately classifying various crop types. The Xception model's strength lies in its unique architecture constructed using depthwise separable convolutions. This design enables the model to learn more intricate and nuanced features from the images, essential in accurately differentiating between various crop types. Despite being highly performant, the model is still relatively efficient for its size of 80.03 MB and a prediction time of 0.0633 seconds. These make it practical for real-world applications where accuracy and efficiency are essential. While ResNet152V2 and InceptionResNetV2 also showed excellent performance, with test accuracies of 98\% and 93\%, respectively, the Xception model's balance of high accuracy, efficiency, and generalization capability makes it the most suitable model for this specific task of crop classification among the transfer learning models evaluated. This emphasizes the importance of selecting a model based on its raw performance metrics, as well as its architecture and suitability for the intended application.

\subsection{Fourth Approach: State-of-the-Art Foundation Models}
In the fourth approach to crop classification, the foundation models, YOLOv8 and DINOv2, have been implemented, representing cutting-edge developments in deep learning and computer vision. These models demonstrate advanced neural networks' evolving capabilities and potential in specialized tasks like crop classification.

\begin{table}[ht]
\centering
\caption{Performance metrics of YOLOv8}
\label{tab:yolov8_metrics}
\resizebox{\textwidth}{!}{%
\begin{tabular}{@{}lllll@{}}
\toprule
\textbf{Model} & \textbf{Top-1 Accuracy} & \textbf{Top-5 Accuracy} & \textbf{Model Size (MB)} & \textbf{Time (sec)} \\ 
\midrule
YOLOv8 & 99.5\% & 100\% & 3.0 & 0.126 \\
\bottomrule
\end{tabular}%
}
\end{table}

\begin{figure}[h]
  \centering
   \includegraphics[width=12cm, height=10cm]{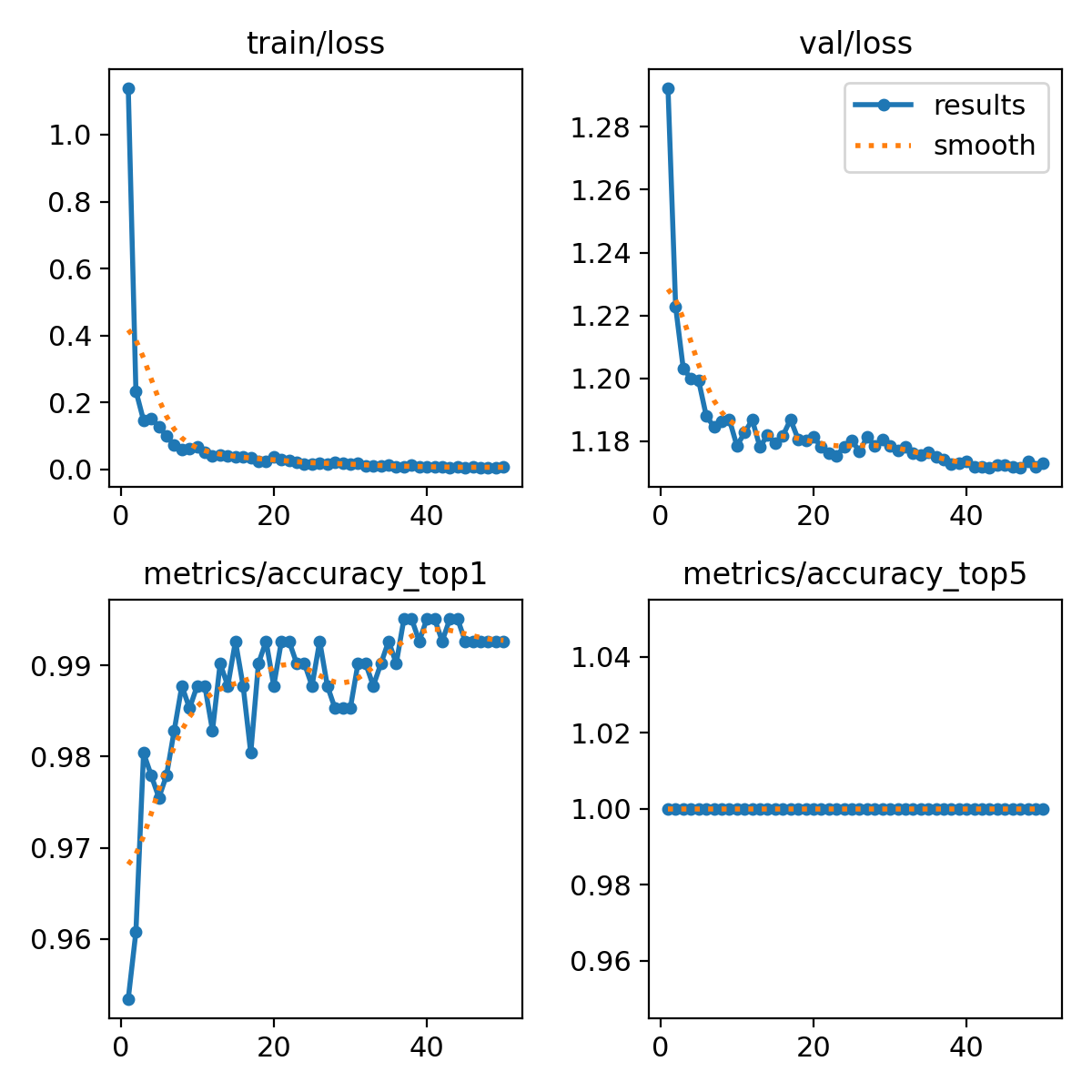}
   \caption{Metrics results for yolov8 used as classification model}
   \label{fig:111}
\end{figure}
YOLOv8, the latest iteration in the YOLO series, is renowned for its object detection capabilities but can also be used for classification, as mentioned. In the study, YOLOv8 shows remarkable effectiveness, achieving a top-1 accuracy of 99.5\% and a top-5 accuracy of 100\%. This level of precision indicates its robustness in accurately identifying and classifying different crop types. The model prediction time is 0.126 seconds, highlighting its prediction efficiency, a crucial factor for time-sensitive applications. Furthermore, YOLOv8's model size is only 3.0 MB, allowing easy deployment in storage and memory-constrained environments. Its high accuracy, speed, and compact size make it an optimal solution for crop classification tasks that require real-time and precise performance and efficiency. The results are also shown in the table \ref{fig:111}, which has all learning curves of different metrics like training loss, validation loss, accuracy for top-1 and accuracy for top-5.

\begin{figure}[h]
  \centering
   \includegraphics[width=12cm, height=12cm]{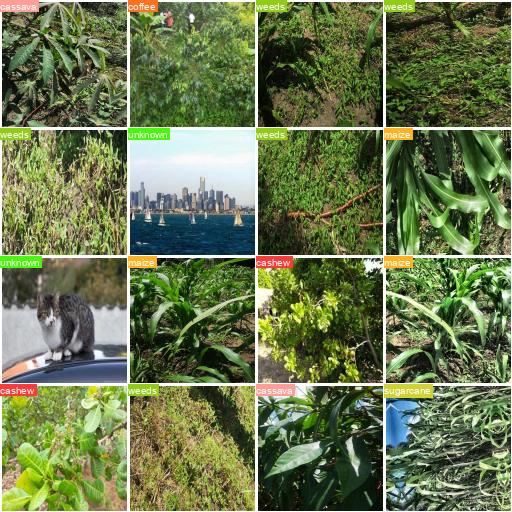}
   \caption{Samples of images from the test classified using Yolov8}
   \label{fig:112}
\end{figure}

Figure \ref{fig:112} shows some samples of images in the test set and how they have been classified using yolov8 as a classifier. We can see from the image that all images were classified correctly, and none were classified incorrectly. 


\begin{table}[ht]
\centering
\caption{Performance metrics of different feature extractors with KNN and SVM models}
\label{tab:metrics}
\resizebox{\textwidth}{!}{%
\begin{tabular}{@{}lllllllllll@{}}
\toprule
\multicolumn{1}{c}{\textbf{Feature Extractors}} & \multicolumn{1}{c}{\textbf{Models}} & \multicolumn{1}{c}{\textbf{Tr. Acc}} & \multicolumn{1}{c}{\textbf{Test Acc}} & \multicolumn{1}{c}{\textbf{Prediction}} & \multicolumn{1}{c}{\textbf{Recall}} & \multicolumn{1}{c}{\textbf{F1-Score}} & \multicolumn{1}{c}{\textbf{Size/MB}} & \multicolumn{1}{c}{\textbf{Time/seconds}} \\ 
\midrule
\multicolumn{1}{l}{DINOv2} & KNN & 98.99\% & 97.73\% & 98\% & 98\% & 98\% & 6.713 & 0.1832 \\
\multicolumn{1}{l}{} & SVM & 99.96\% & 99.65\% & 99\% & 99\% & 99\% & 2.385 & 0.0698 \\ \bottomrule
\end{tabular}%
}
\end{table}

DINOv2 employs a self-supervised Vision Transformer model for feature extraction, which is then used with classical machine learning models such as KNN and SVM. The high test accuracies achieved with DINOv2 97.73\% with KNN and 99.65\% with SVM are a testament to the efficacy of transformers in capturing complex features from images. This combination showcases the potential of advanced feature extraction with traditional classifiers to enhance the performance of machine learning algorithms. The precision, recall, and F1 scores for DINOv2 demonstrate its ability to provide accurate and dependable classifications, achieving 98\% accuracy for KNN and 99\% for SVM. The sizes and processing times of the models, especially the SVM model, show an effective utilization of computational resources. They strike a balance between accuracy and operational demands. The prediction time for KNN is 0.1832 seconds, while for SVM, it is 0.0698 seconds. The model size for KNN is 6.713 MB, while for SVM, it is 2.385 MB. Although this method captures specific aspects of crop data, it still has limitations similar to the first approach, making it imperfect for deployment.

Another approach tried is to use a foundation model to train crop detection without labelling. Autodistill was used as a new and innovative method for creating computer vision models without labelling training data. This approach could revolutionize how enterprise AI applications and computer vision models are developed. Autodistill uses large foundation models to transfer their extensive knowledge to more compact models. The process begins by combining all crop images into a single folder, bypassing the need for individual labels. This is where Autodistill's unique capabilities come into play. Autodistill offers a Python method to define prompts or labels, known as "Ontologies", and integrates Grounded SAM. This plays a crucial role in identifying the desired categories, in this case, six distinct classes of crops. Grounded SAM skillfully combines Sharpness-Aware Minimization (SAM) with segmentation techniques to delineate specific parts of an image and Grounding DINO for labelling the contents within these segments. Once these initial steps are completed, the focus shifts to training a new crop identification model using the robust YOLOv8 architecture. This phase is crucial as it transforms the conceptual framework provided by Autodistill into a practical, efficient tool for identifying various crop classes. 

\begin{figure}[h]
    \centering
    \begin{minipage}{.5\textwidth}
        \centering
        \includegraphics[width=8cm, height=6cm]{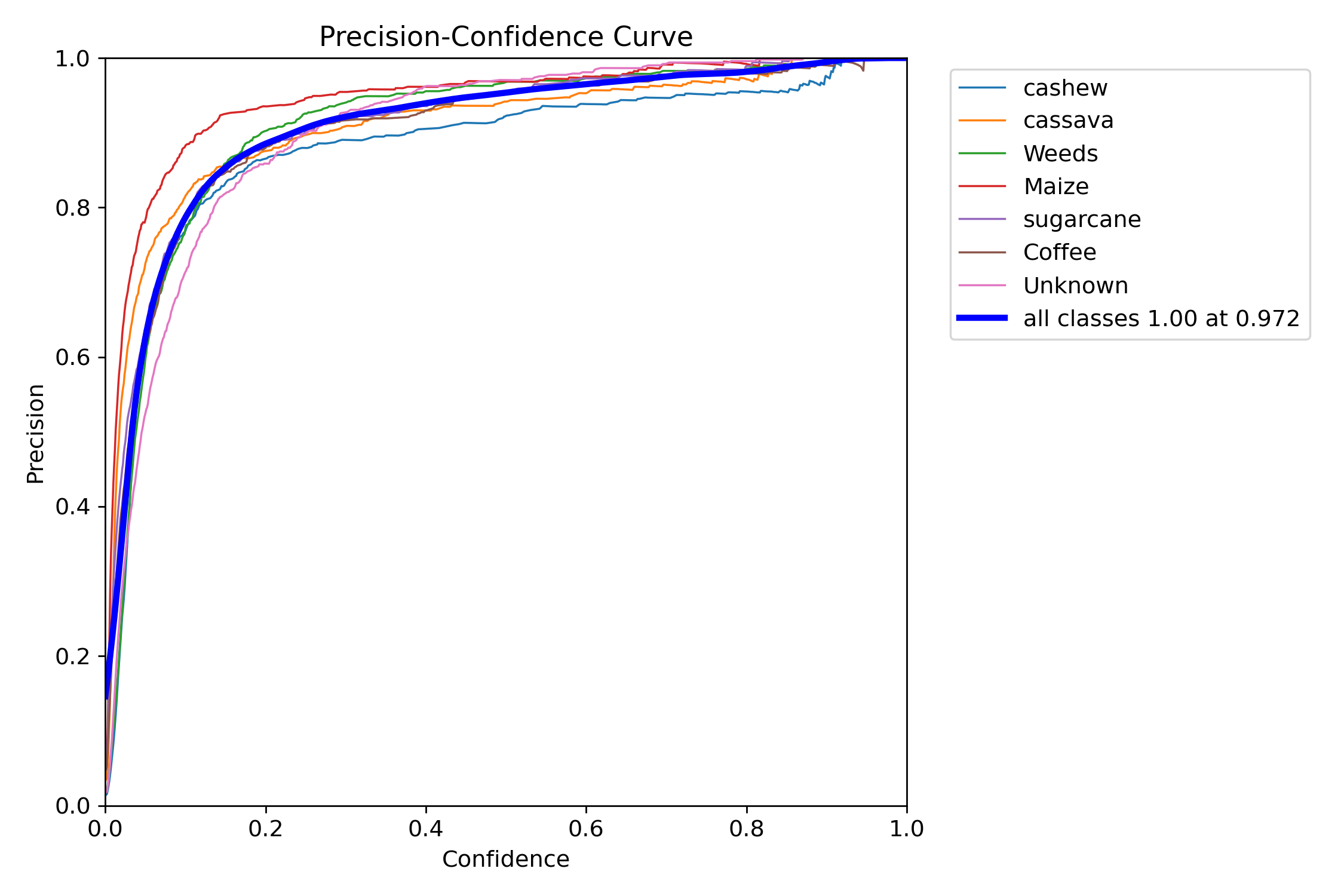}
        \caption{Precison-Confidence Curve of Trained YOLOv8 with Autodistil}
        \label{fig:31}
    \end{minipage}%
    \begin{minipage}{.5\textwidth}
        \centering
        \includegraphics[width=8cm, height=6cm]{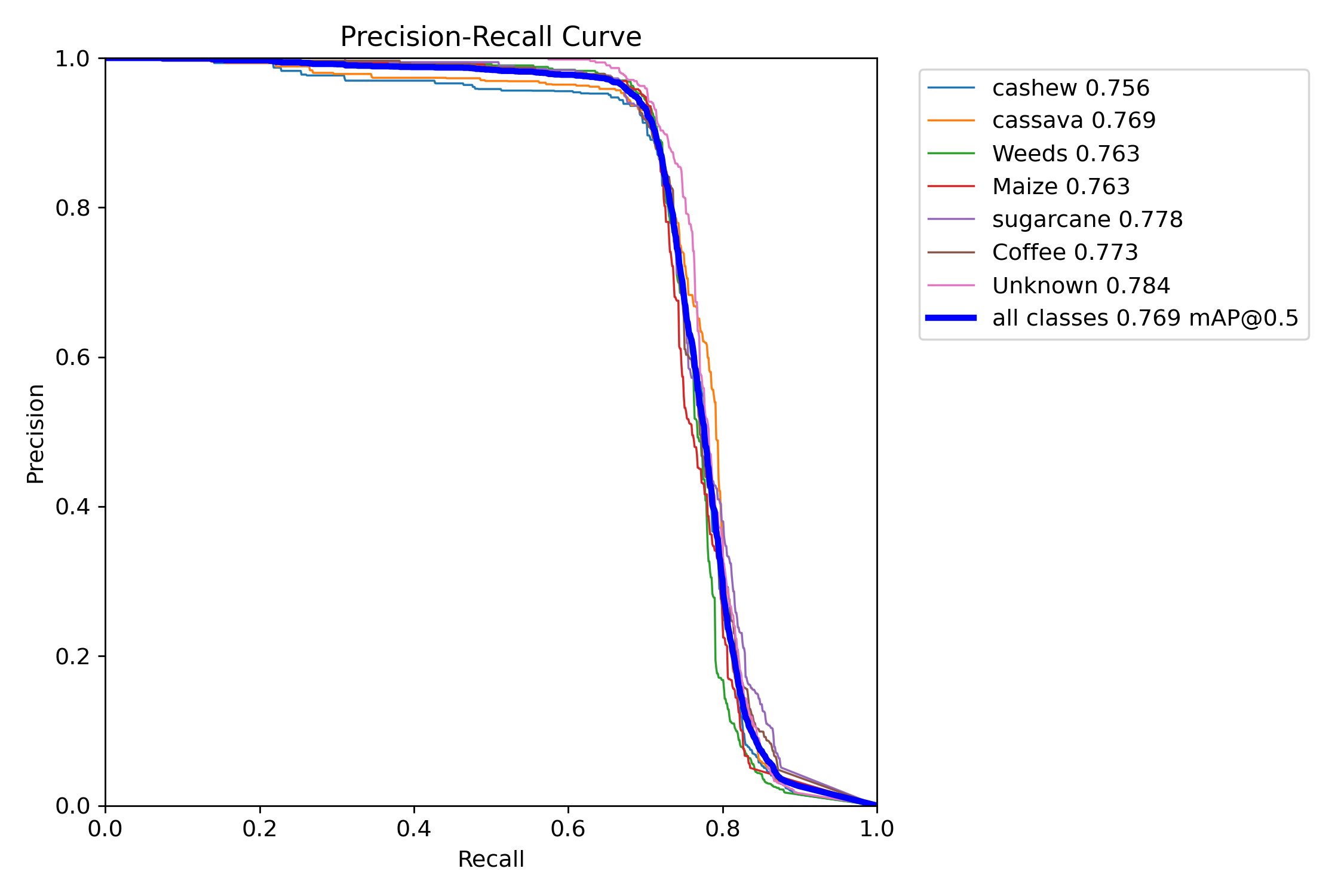}
        \caption{Precison-Recall Curve of Trained YOLOv8 with Autodistil}
        \label{fig:32}
    \end{minipage}
\end{figure}

Figure \ref{fig:31} represents the Precision-Confidence Curve, which plots the model's precision against varying confidence thresholds. Each line corresponds to one of the six crop classes recognized by the model, namely cashew, cassava, maize, weeds, sugarcane, and coffee, with an additional line representing an 'Unknown' class and one for 'all classes' at a threshold of 0.9. A high precision level at a high confidence threshold indicates that the model is highly accurate when confident about its predictions. The convergence of all classes towards the high-precision end of the curve as confidence increases demonstrates the model's ability to maintain accuracy across different crop types. The additional 'all classes' line at 0.9 suggests that the model achieves near-perfect precision across all classes when it reaches a 90\% confidence threshold.

Figure \ref{fig:32} represents the Precision-Recall Curve, a common performance metric for object detection tasks where the prevalence of classes is imbalanced. Each curve represents one of the crop classes, plotting the precision of the model against its recall. Precision measures the accuracy of the positive predictions, while recall measures the model's ability to find all the positive samples. The area under these curves (AUC) provides an aggregate performance measure across all thresholding levels. The mAP@0.5 score (mean Average Precision at 0.5 thresholds) for each class is indicated in the legend, summarizing the overall precision and recall performance. A high mAP@0.5 score signifies a precise and robust model in detecting the presence of crops across various classes. The 'all classes' line reports a mAP@0.5 of 0.769, indicating that the model demonstrates a good balance between precision and recall across all crop classes at the specified threshold on average.

These 2 curves suggest that the train yolov8 without labels has a robust predictive performance, with high precision at high confidence levels and commendable average precision across different recall levels.

\subsection{General Comparison}
In summarizing the four distinct approaches to crop classification: traditional machine learning with feature extraction, custom and established deep learning architectures, transfer learning models, and foundation models, a comprehensive understanding emerges, highlighting the strengths and suitability of each method for specific scenarios.

\begin{itemize}
    \item \textbf{Traditional Machine Learning with Feature Extraction:} The combination of Color Histogram and SVM emerged as the best in this category, showing high effectiveness with a test accuracy of 93\% and precision, recall, and F1-scores above 90\%. It is notably efficient regarding computational resources, making it suitable for environments with limited processing power. However, relying on manually selected features might limit its capability for deployment.

    \item \textbf{Custom and Established Deep Learning Architectures:} The custom CNN in this approach displayed remarkable performance, achieving training accuracy of 96\% and test accuracy of 92\%. In contrast, established architectures AlexNet showed lower effectiveness in this specific task. The custom CNN’s tailored design enables it to capture nuanced features of crops effectively but at the cost of larger model sizes and higher computational demands, in addition to the time used for fine-tuning the parameters to get the best ones.

    \item \textbf{Transfer Learning Models:} Xception stands out with its exceptional training accuracy of 99\% and validation accuracy of 99\% and high test accuracy of 98\%. It also demonstrated high precision, recall, and F1-scores at 98\%. Despite its larger model size of 80.03 MB and computational requirements, Xception offers a robust balance of high accuracy and efficiency. It is suitable for deployment in various applications where precision in image classification is key, and the size is relatively small compared to other models, except for MobilenetV3. However, it doesn't perform well in terms of accuracy.

    \item \textbf{Foundation Models:} YOLOv8 showcased its superiority with a top-1 accuracy of 99.5\% and a top-5 accuracy of 100\%, a prediction time of 0.126 seconds and a compact model size of 3.0 MB. DINOv2, when combined with KNN, achieved a test accuracy of 97.73\%, and with SVM, it reached an even higher test accuracy of 99.65\%. Both YOLOv8 and DINOv2 demonstrate the advances in neural networks, offering high accuracy and efficiency, crucial for deployment in varied agricultural scenarios.

\end{itemize}

Finally, YOLOv8, from the foundation models category, emerges as the most effective for crop classification tasks in determining the best model for practical implementation. Its outstanding high accuracy, speed, and efficiency balance make it highly practical for real-time applications. Alternatively, Xception from the transfer learning models also presents as a strong contender, especially in scenarios where detailed image analysis is critical, and there is flexibility regarding computational resources. The choice between YOLOv8 and Xception should be based on the application's specific requirements and the deployment environment, with YOLOv8 being ideal for high-speed and accurate object detection and Xception excelling in detailed image classification tasks.

\begin{figure}[h]
    \centering
    \begin{minipage}{.5\textwidth}
        \centering
        \includegraphics[width=8cm, height=6cm]{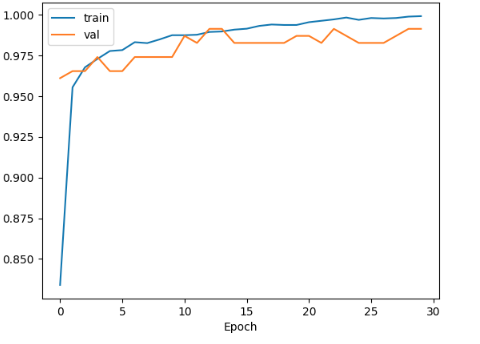}
        \caption{Training and validations accuracy curves for Xception model}
        \label{fig:10}
    \end{minipage}%
    \begin{minipage}{.5\textwidth}
        \centering
        \includegraphics[width=8cm, height=6cm]{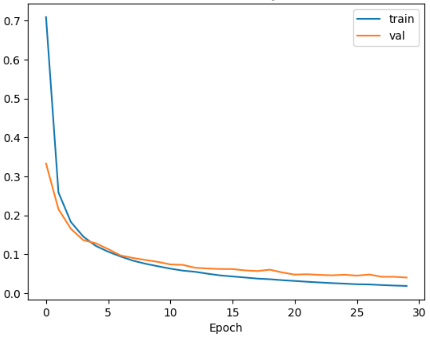}
        \caption{Training and validations loss curves for Xception model}
        \label{fig:11}
    \end{minipage}
\end{figure}

The graph illustrated in Figure 5 demonstrates the accuracy of the Xception model over 30 epochs on both the training and validation datasets. The blue line represents the model's training accuracy, while the orange line represents the validation accuracy. The accuracy of both lines increases rapidly, achieving 99\% accuracy and indicating that the model quickly adapts to the training data and performs consistently on the validation data. Additionally, the training process shows that the curves are closely aligned and parallel, indicating that the model is well-balanced. The model has effectively learned from the training data and can generalize to new and unseen data with minimal overfitting.

In Figure 6, we can see the loss curves for the Xception model. These curves reflect how the model's error rate is optimized across epochs. The blue line represents the training loss, which starts with a sharp decline from 1.0 to 0.01. Similarly, the orange line represents the validation loss, sharply declining from 0.5 to 0.04. As the model continues to learn, the error rate decreases until it reaches a stable level. The training and validation loss curves level off at this point, indicating that the model has converged and no longer improves. This behaviour indicates a well-tuned model that learns to distinguish between different features without simply memorizing the training data.

\begin{figure}[h]
  \centering
   \includegraphics[width=10cm, height=8cm]{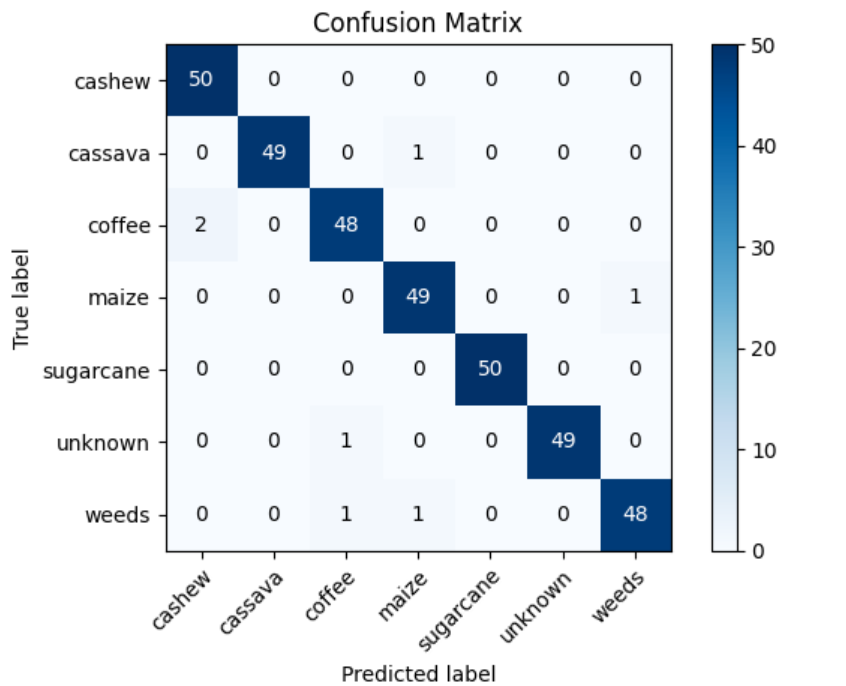}
   \caption{Confusion matrix for the Xception model}
   \label{fig:ke1}
\end{figure}

On the confusion matrix shown in \ref{fig:ke1} of the test set, there were 50 images for every class of cashew, cassava, coffee, maize, sugarcane, unknown, and weeds. Across all classes, there were 350 images, of which only 7 were misclassified. This means the model is highly accurate, with most of its predictions correctly matching their true labels. Overall, the matrix reveals that the Xception model is highly effective.

XAI remains imperative despite the high accuracy and promising training/validation curves presented by the Xception model and a largely convincing confusion matrix. XAI can provide transparency into a model's decision-making process, allowing the user to validate the model's performance and understand the features and patterns used to make predictions. Furthermore, XAI helps detect biases in datasets and models, ensuring fair, accountable and ethical decisions. 

\subsection{Explainable AI}

It is impressive to achieve high accuracy in AI models. However, more is needed for practical applications. These models have a 'black-box' nature that conceals the reasoning behind their predictions. Therefore, it is necessary to use XAI to make these models more transparent and understandable. As previously discussed, XAI is crucial for ensuring transparency and ethical integrity in the real-world deployment of AI. XAI allows for understanding how these models arrive at their decisions, which is essential for building trust. XAI techniques such as LIME, SHAP, and Grad-CAM were applied to improve the interpretability of all models. This journal focuses on the XAI of the Xception model, which is considered the best model for this research. Although all models showed impressive capabilities for the task of crop classification, the journal's detailed XAI analysis of the Xception model will provide a deeper understanding of its inner workings. Focusing XAI efforts on the Xception model is pivotal for enhancing trust and transparency among users, refining the model's accuracy, ensuring regulatory compliance, and facilitating inclusive engagement. Additionally, all users can better understand and trust the AI by shedding light on the model's decision-making. This will increase the model's trustworthiness and acceptability and ensure that its decision-making process aligns with the broader goals of fairness, transparency, and ethical responsibility. 

The XAI tools used are indispensable, ensuring that an equal measure of clarity and accountability matches the model's high performance. LIME provides a local approximation of complex models to explain their predictions, shedding light on the role of individual features. SHAP provides a comprehensive understanding of model behaviour by quantifying the contribution of each feature to the prediction. Furthermore, Grad-CAM uses gradient information to highlight the most influential regions in input images, which is especially useful for convolutional neural networks. Collectively, these tools provide transparency into the model's decision-making process, making it more accountable and easy to understand.

\begin{figure}[h]
  \centering
   \includegraphics[width=15cm, height=6cm]{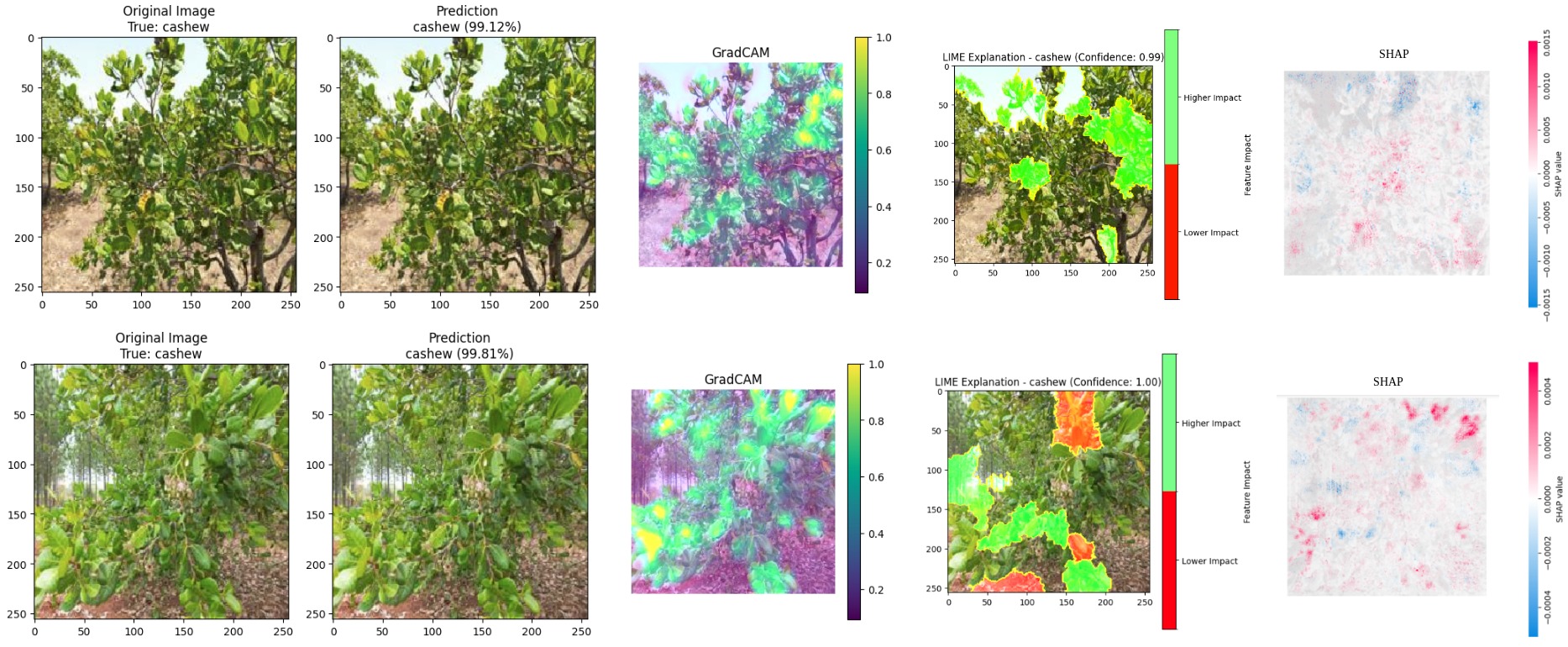}
   \caption{KNN with SIFT confusion matrix on the test data}
   \label{m1}
\end{figure}

\begin{figure}[h]
  \centering
   \includegraphics[width=15cm, height=6cm]{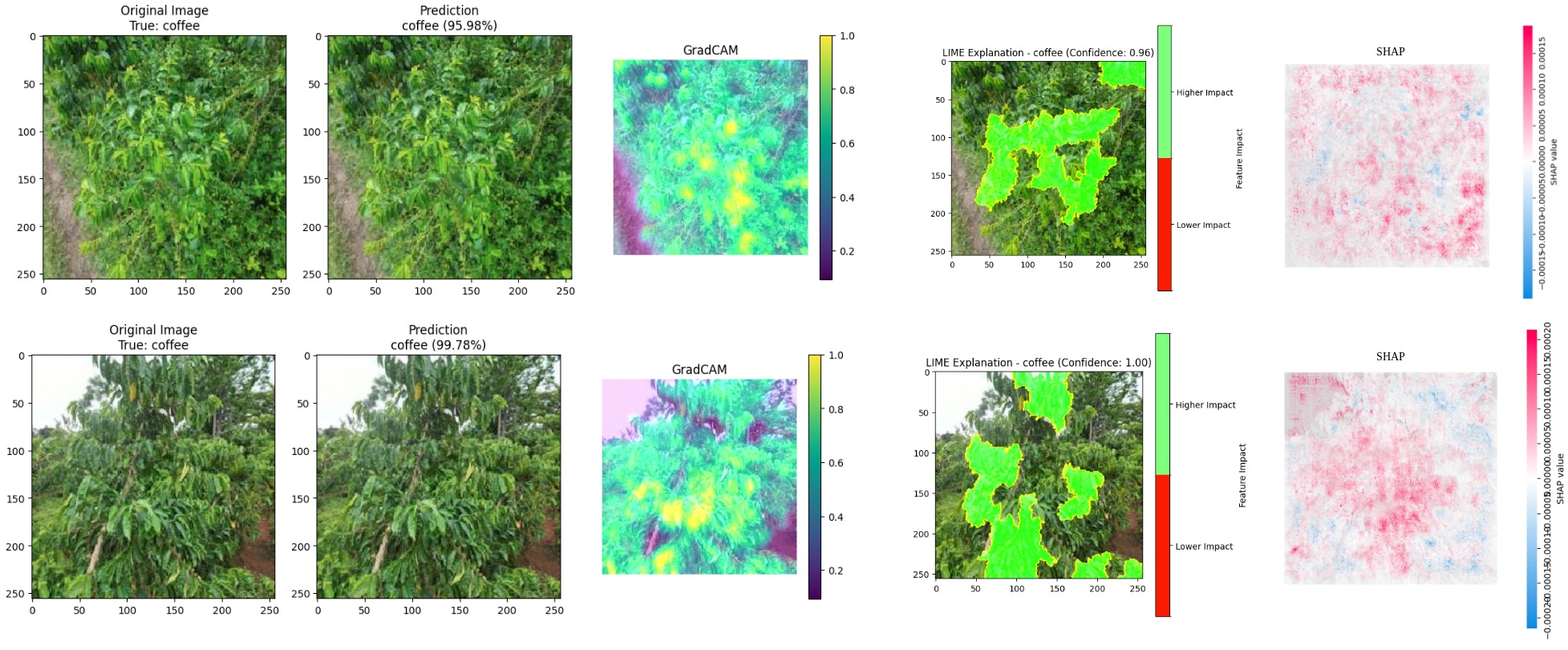}
   \caption{KNN confusion matrix (Color histogram)}
   \label{m2}
\end{figure}

\begin{figure}[h]
  \centering
   \includegraphics[width=15cm, height=6cm]{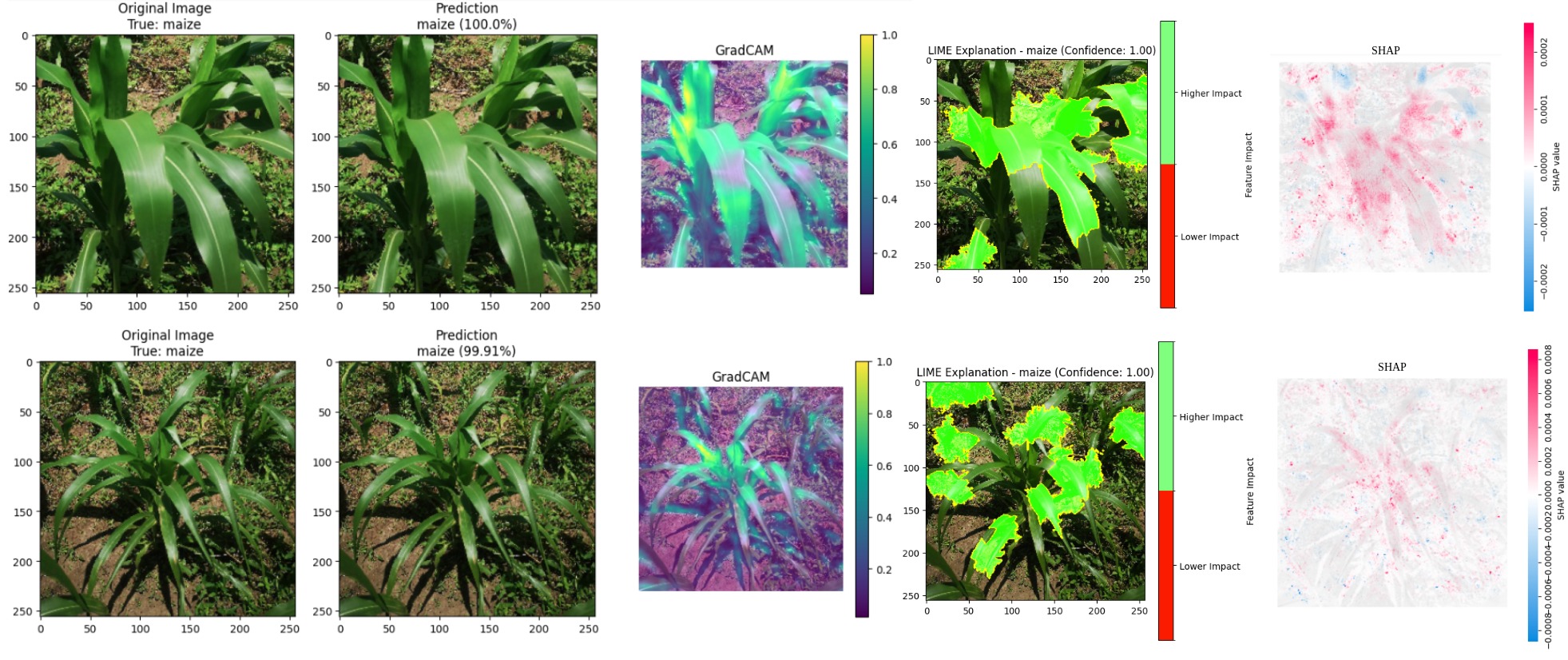}
   \caption{CNN confusion matrix}
   \label{m3}
\end{figure}

The series of images below provide explainability and interpretability of the Xception model. XAI tools generate them, and their corresponding discussions are added. All images are arranged in the order of cashew, coffee, cassava, sugarcane, unknown images, and infield weeds. The images have been explained using GradCAM, Lime, and Shap, and the corresponding figures are identified as \ref{m1}, \ref{m2}, \ref{m3}, \ref{m4}, \ref{m5}, and \ref{m6}, respectively. Additionally, each figure has ten images in one except the weeds in the infield, which have five images. The order of the images is consistent throughout the figures, with the two left-most images being selected from the test set (two images were considered to provide a clear understanding), followed by the two predicted images using the Xception model having the confidence of prediction on top of it, the GradCAM Heatmaps, the LIME Superpixel Importance, and finally, the SHAP SHAP Value Distribution.

\begin{itemize}
    \item \textbf{Grad-CAM Heatmaps:} Grad-CAM uses gradient information from the final convolutional layer of the Xception model to generate a localization map highlighting important regions for predicting the class. Heatmaps use a range of colours to display the gradient tensors' weights. Warmer colours like yellow and green indicate higher weights and a more significant impact on the model's output. On the other hand, cooler colours like purple correspond to lower weights and lesser influence. The heatmap colours are normalized for better visualization, with yellow representing the highest weights and purple the lowest.

    \item \textbf{LIME Superpixel Importance:} LIME uses a model-agnostic approach to interpret images by observing the effect on the output from perturbing the input image. It identifies coherent regions, called superpixels, within the image and evaluates their contribution to the model's prediction. In the results, LIME attributes a confidence score to each superpixel, depicted in varying shades. Brightly coloured green superpixels contribute positively to the model's confidence in its prediction, while darker red shades have a neutral or negative contribution. 

    \item \textbf{SHAP Value Distribution:} SHAP values measure each feature's impact on a model's prediction. Each pixel's SHAP value indicates how much it influences the model output from the base value (the model's output if no features were present) when included in the prediction. The colour gradient on the image indicates the distribution of SHAP values. The red spots show the features that increase the model's output (positive SHAP values), while the blue spots show the features that decrease the output (negative SHAP values). This distribution gives an overview of the significance of the features, and the magnitude of the SHAP values reflects the strength of each feature's impact on the prediction.

\end{itemize}

In Figure \ref{m1}, we can see cashew crop predictions and explainability. The Xception model predicted cashew correctly in both images with 99.12\% and 99.81\% confidence. The GradCAM highlighted regions in yellow and green around cashew leaves, indicating a high impact. The ground, trees, and sky in the background were highlighted with purple colour, indicating low impact. LIME also identified specific regions or superpixels with high impact in green and low impact in red, but not as well as GradCAM. The Shap explainer also highlighted features around the cashew leaves.

In Figure \ref{m2}, we can see coffee crop predictions and explainability. The two coffee sample images were correctly predicted with 95.98\% and 99.78\% confidence. Their Grad-CAM heatmaps illuminate vital areas contributing to the model's classification with varying intensity for the coffee images. The highlighted regions, predominantly around the foliage and the clusters of coffee cherries, align with agronomic knowledge that these characteristics are distinctive for coffee crops. The warmer colours on the heatmaps represent the intensity with which the model confidently identifies coffee within specific regions. The areas highlighted in purple, such as the sky in the second image, indicate a negative impact on the region. The LIME outlined superpixels on the coffee images, bright and prominent against the backdrop, signify the model's reliance on these patches for its decision. The high confidence levels of 0.96 and 1.00 reinforce the model's classification capability. The identified features align solidly with the coffee class. Scattered positive contributions throughout the coffee crop images suggest the model considers various features when assessing in SHAP.

In Figure \ref{m3}, we can see maize crop predictions and explainability. The two maize sample images were correctly predicted with 100.0\% and 99.91\% confidence. The Grad-CAM heatmaps guide the areas of focus within the maize imagery that the model uses to classify the images. The heatmaps display warmer colours, primarily around the edges and central veins of the maize leaves, corresponding to higher weights in the gradient tensors. This suggests that these areas have a significant influence on the classification output. The heatmaps confirm that the model mostly focuses on the structural attributes of the maize plants, which are critical identifiers for the maize class. For LIME, the images of maize show a large, connected precision agriculture domain, machine learning models' interpretability, and that the model is highly confident in its prediction. The highlighted regions correspond to the green hues and textures of the maize leaves, emphasizing their significance in the model's predictive process. For SHAP, the values are distributed across the maize imagery, providing a nuanced picture of each pixel's contribution to the model's decision. The image shows scattered positive (red) and negative (blue) SHAP values, with a concentration of positive values on the maize leaves.

\subsubsection{\textbf{Sugarcane}}

\begin{figure}[h]
  \centering
   \includegraphics[width=15cm, height=6cm]{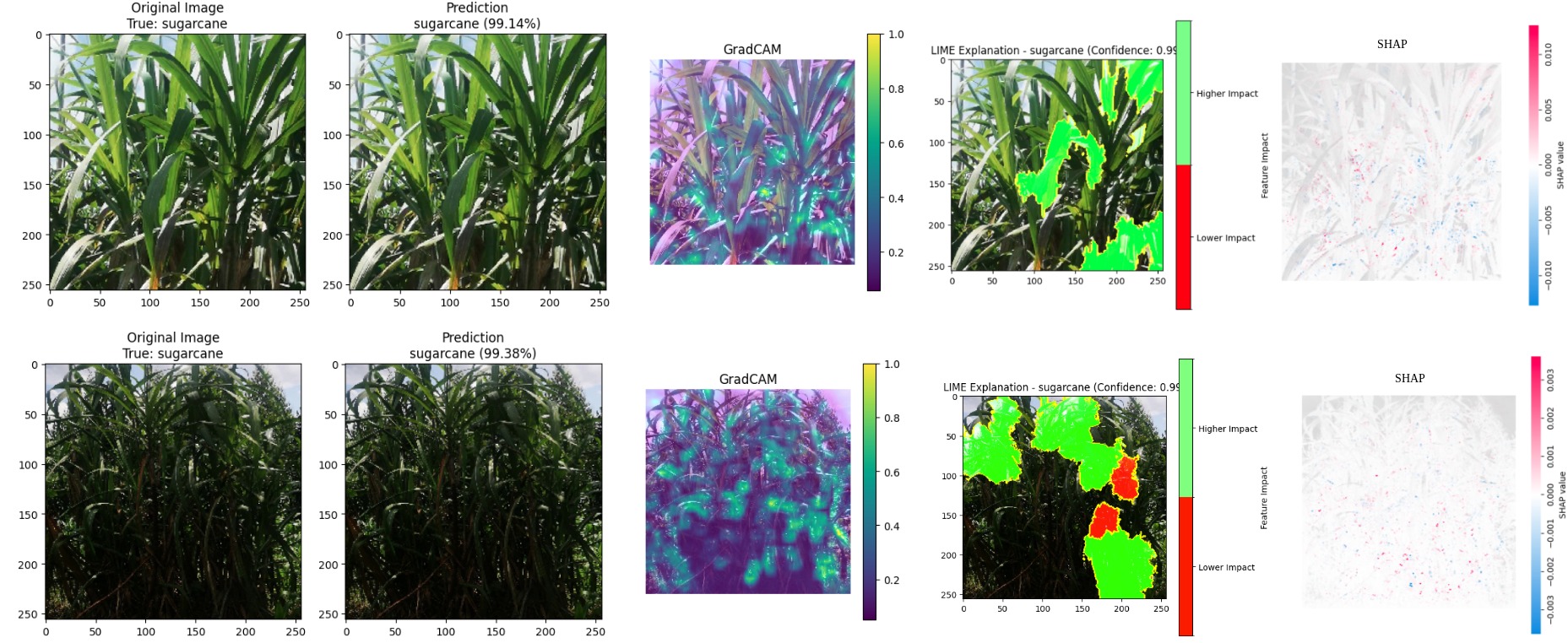}
   \caption{KNN confusion matrix (ORB)}
   \label{m4}
\end{figure}

In Figure \ref{m4}, we can see sugarcane crop predictions and explainability. The two sugarcane sample images were correctly predicted with 99.14\% and 99.38\% confidence. The Grad-CAM The highlighted regions where the heatmaps display warmer colours, primarily around the edges and central veins of the maize leaves, which correspond to higher weights in the gradient tensors and suggest a significant influence on the classification output. The heatmaps confirm that the model primarily focuses on the structural attributes of maize plants, which are critical identifiers for the maize class. For the LIME, the maize images exhibit large contiguous regions highlighted with a confidence score 1.00, reflecting the model's absolute certainty in its predictions. These bright, highlighted areas correspond to the maize leaves' distinctive green hues and textures, underscoring their importance in the model's predictive process. The SHAP values are distributed across the maize imagery, providing a nuanced picture of each pixel's contribution to the model's decision. The positive and negative SHAP values are scattered across the image, with a concentration of positive values along the maize leaves. This distribution indicates the model's comprehensive assessment, integrating a broad spectrum of features within the image, affirming the crop's classification.

\begin{figure}[h]
  \centering
   \includegraphics[width=15cm, height=6cm]{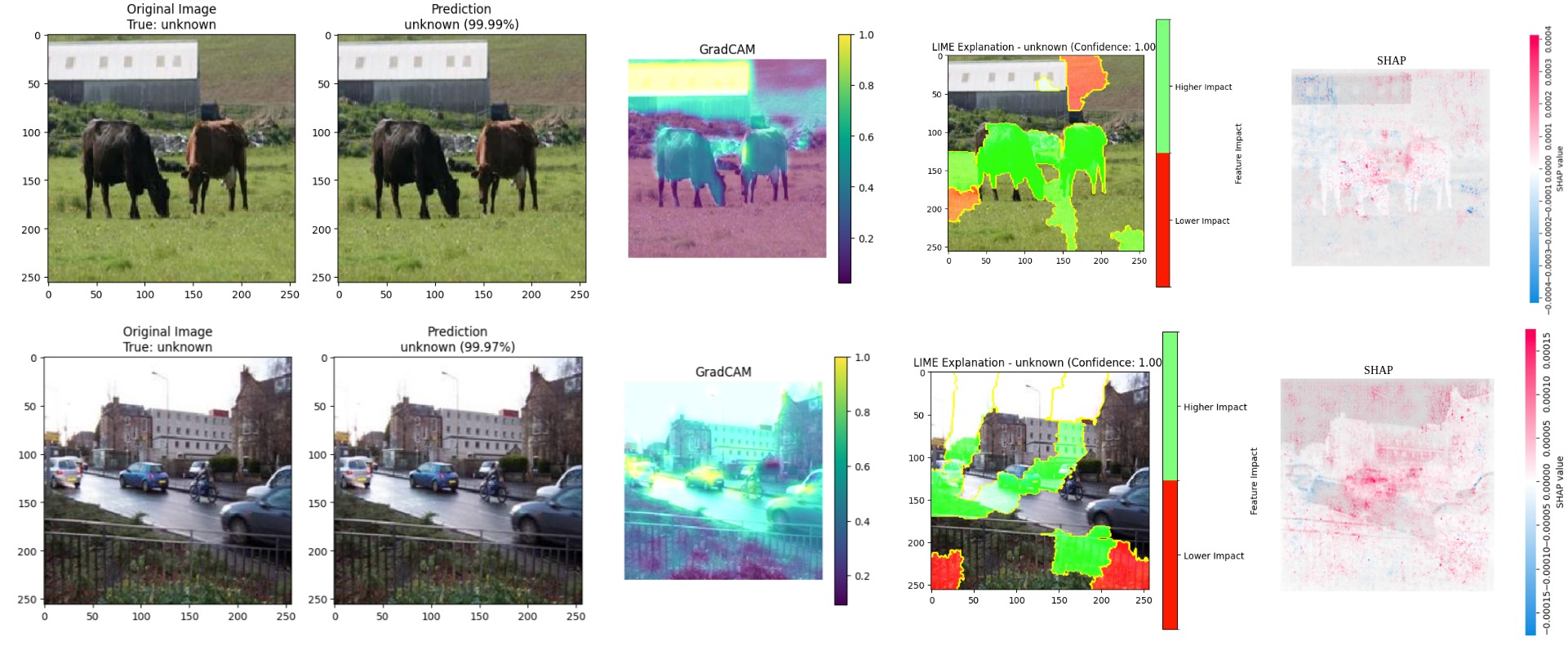}
   \caption{KNN confusion matrix (Colorhistogram)}
   \label{m5}
\end{figure}

In Figure \ref{m5}, we can see unknown images and their explainability. The two unknown images were correctly predicted with 99.99\% and 99.97\% confidence. The GradCAM heatmaps for the unknown images do not focus on a specific pattern within the data, as would be expected for known classes. Instead, the highlighted areas are dispersed across the image, suggesting the absence of distinct, class-specific features that the model recognizes. This dispersion of attention indicates the model's attempt to find familiar patterns and assign the unknown label upon failing. Elements such as cows, houses, and cars were highlighted as positively impacting the unknown class. However, in image two, some grass appeared purple because it can resemble weeds in the infield. For LIME, it outlines the regions within the images that influenced the model's unknown prediction. The highlighted bright-colour superpixels are less consistent and more scattered than those in known class images. The confidence scores, approaching 1.00, reflect the model's high certainty without known class indicators, leading to the unknown classification. For SHAP, the values across the unknown images present a stark pattern different from known classes. Instead of pinpointing decisive features contributing to a specific class, the SHAP values appear more randomized and less focused, aligning with the model's image classification as unknown.

\begin{figure}[h]
  \centering
   \includegraphics[width=15cm, height=3.5cm]{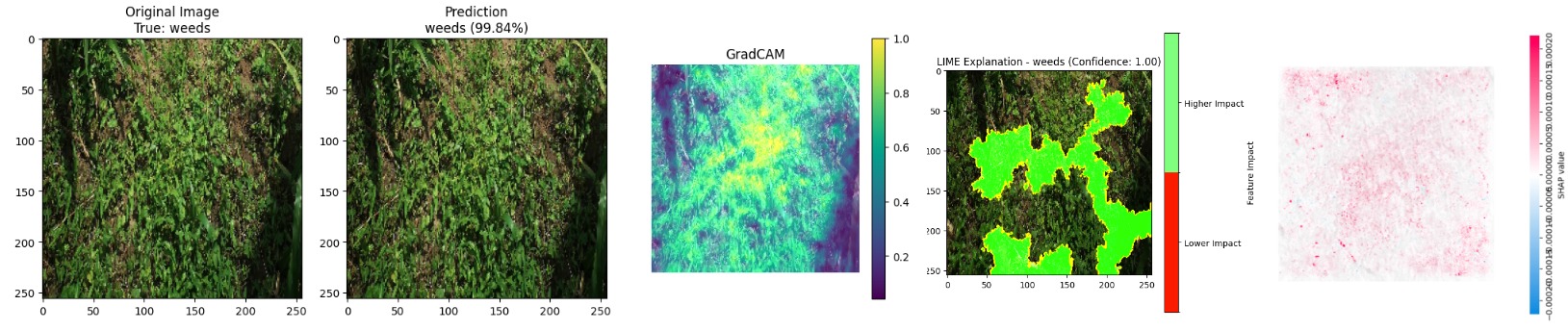}
   \caption{KNN confusion matrix (Colorhistogram)}
   \label{m6}
\end{figure}

Lastly, in Figure \ref{m6}, we can see the image and explainability of weeds in the field. The image was predicted with 99.84\%  confidence. The Grad-CAM yields a heatmap that underscores the regions within the image most influential to the model prediction. The vibrant areas in the Grad-CAM visualizations suggest that the model primarily focuses on the textural patterns and colour variations characteristic of weed foliage, which often differ from the crops they infest. This heatmap provides a visual confirmation that the model focus aligns with recognizable weed attributes, which is crucial for precision agriculture and automated weed management systems. The LIME breaks down the prediction by highlighting the superpixels in the image most instrumental in classifying weeds. The highlighted bright-coloured regions correlate with a high confidence level (1.00), suggesting that the model certainty stems from specific features associated with the weeds. These highlighted superpixels, which likely represent the unique textural patterns of the weeds against the soil or crop background, validate the modality to distinguish between crops and invasive species. SHAP values assign a contribution score to each pixel, showing the positive or negative influence each has on the model output. The subtle patterns of SHAP values across the image reinforce the notion that the model is assessing a comprehensive array of features, and the predominance of positive values in areas corresponding to the weeds confirms their impact on the classification decision.

Utilizing these tools offers a comprehensive understanding of a model's predictive focus. Grad-CAM heatmaps indicate specific regions of interest within an image. LIME assess and ranks individual superpixels based on their impact, while SHAP values provide a detailed contribution map across the entire image using features. Collectively, these techniques affirm the model's high accuracy and provide transparency into its operational mechanics. This transparency is paramount for the scientific community's trust, ensuring the model's predictions are grounded in observable and relevant features within the image data.

\section{Conclusion}
During this study, we systematically compared four approaches to crop classification: traditional machine learning, deep learning architectures, transfer learning models, and state-of-the-art foundation models. Our investigation revealed different methodologies' nuanced capabilities and limitations in identifying crops and anomalies in agricultural imagery. All code and models are saved on GitHub; \href{https://github.com/sudimuk2017/Interpretable-Machine-Learning-Approaches-for-Crop-Classification/tree/main}{here is the link}.

The traditional machine learning approach, particularly the SVM combined with Color Histogram feature extraction, demonstrated a balance between performance and efficiency with 96\% training accuracy, 93\% test accuracy, 94\% prediction, recall, and F1 score, a 5.88 MB model size, and 0.0027 seconds for prediction. However, for Deep Learning models, the custom-designed CNN and Xception models truly showcased the potential of deep learning, achieving remarkable accuracy and generalization in classification tasks. The custom-designed CNN model achieved 96\% training accuracy, 94\% validation accuracy, 0.08 training loss, 0.40 validation loss, 92\% test accuracy, 92\% precision, 92\% recall, 92\% F1 score, 85.08 model size, and 0.0755 second prediction time. The Xception model, when further enhanced with transfer learning techniques, stood out for its exceptional performance. It provided robust accuracy and adaptability, making it an excellent choice for crop classification. It got a training accuracy of 99\%, validation accuracy of 99\%, training loss of 0.01, validation loss of 0.04, test accuracy of 98\%, prediction of 98\%, recall of 98\%, f1 score of 98\%, a model size of 80.03 MB, and a prediction time of 0.0633 seconds. The foundation models, notably YOLOv8, brought a new dimension to the task with their rapid processing capabilities and high predictive accuracy. It achieved 99.5\% top-1 accuracy, 100\% top-5 accuracy, 3.0MB model size, and 0.126 sec prediction time. The integration of DINOv2 highlighted the synergy between advanced neural network architectures and traditional machine learning classifiers, offering an innovative approach to feature extraction and subsequent classification. DINOv2 achieved 99.96\% training accuracy, 99.65\% test accuracy, 99\% prediction, recall, and f1 score. The model size is 2.385MB, and the prediction time is 0.0698 seconds.

We gained valuable insights into how the models make decisions by implementing Explainable AI techniques, including Grad-CAM, LIME, and SHAP. This was essential for validating the accuracy of the model's predictions, ensuring transparency, building trust, and helping understand which models are perfect for real-world agricultural scenarios. The XAI analysis confirmed that the high performance of models like Xception was due to their focus on relevant features, aligning with domain expertise in crop identification. It has also demonstrated the indispensable role of XAI in making AI more interpretable and trustworthy.

As we move towards more AI-driven agricultural practices, the future of this field could involve improving the number of crops used for classification to create a more extensive set of crops. Also, the best model could be deployed in an infield application to assist farmers. This model can be used to estimate the yield of crops such as coffee or cashew after detecting the actual crop. The insights gained from this study will be crucial in enhancing crop management strategies and ultimately leading to more efficient and sustainable farming practices.

\bibliographystyle{ACM-Reference-Format}
\bibliography{main}


\begin{thebibliography}{63}


\ifx \showCODEN    \undefined \def \showCODEN     #1{\unskip}     \fi
\ifx \showDOI      \undefined \def \showDOI       #1{#1}\fi
\ifx \showISBNx    \undefined \def \showISBNx     #1{\unskip}     \fi
\ifx \showISBNxiii \undefined \def \showISBNxiii  #1{\unskip}     \fi
\ifx \showISSN     \undefined \def \showISSN      #1{\unskip}     \fi
\ifx \showLCCN     \undefined \def \showLCCN      #1{\unskip}     \fi
\ifx \shownote     \undefined \def \shownote      #1{#1}          \fi
\ifx \showarticletitle \undefined \def \showarticletitle #1{#1}   \fi
\ifx \showURL      \undefined \def \showURL       {\relax}        \fi
\providecommand\bibfield[2]{#2}
\providecommand\bibinfo[2]{#2}
\providecommand\natexlab[1]{#1}
\providecommand\showeprint[2][]{arXiv:#2}

\bibitem[Abbas et~al\mbox{.}(2023)]%
        {abbas2023towards}
\bibfield{author}{\bibinfo{person}{Adel Abbas}, \bibinfo{person}{Michele Linardi}, \bibinfo{person}{Etienne Vareille}, \bibinfo{person}{Vassillis Christophides}, {and} \bibinfo{person}{Claudia Paris}.} \bibinfo{year}{2023}\natexlab{}.
\newblock \showarticletitle{Towards Explainable AI4EO: An Explainable Deep Learning Approach for Crop Type Mapping using Satellite Images Time Series}. In \bibinfo{booktitle}{\emph{IGARSS 2023-2023 IEEE International Geoscience and Remote Sensing Symposium}}. IEEE, \bibinfo{pages}{1088--1091}.
\newblock


\bibitem[Abdelmoneim et~al\mbox{.}(2019)]%
        {abdelmoneim2019comparative}
\bibfield{author}{\bibinfo{person}{Samar~M Abdelmoneim}, \bibinfo{person}{Mohammed Kayed}, {and} \bibinfo{person}{Shereen~A Taie}.} \bibinfo{year}{2019}\natexlab{}.
\newblock \showarticletitle{A comparative study for feature extraction and classification of images}. In \bibinfo{booktitle}{\emph{2019 6th International Conference on Advanced Control Circuits and Systems (ACCS) \& 2019 5th International Conference on New Paradigms in Electronics \& information Technology (PEIT)}}. IEEE, \bibinfo{pages}{105--110}.
\newblock


\bibitem[Aditi and Dureja(2021)]%
        {aditi2021review}
\bibfield{author}{\bibinfo{person}{Aditi} {and} \bibinfo{person}{Aman Dureja}.} \bibinfo{year}{2021}\natexlab{}.
\newblock \showarticletitle{A review: Image classification and object detection with deep learning}. In \bibinfo{booktitle}{\emph{Applications of Artificial Intelligence in Engineering: Proceedings of First Global Conference on Artificial Intelligence and Applications (GCAIA 2020)}}. Springer, \bibinfo{pages}{69--91}.
\newblock


\bibitem[Ali et~al\mbox{.}(2023)]%
        {ali2023transfer}
\bibfield{author}{\bibinfo{person}{Ahmed~Hussein Ali}, \bibinfo{person}{Mohanad~G Yaseen}, \bibinfo{person}{Mohammad Aljanabi}, {and} \bibinfo{person}{Saad~Abbas Abed}.} \bibinfo{year}{2023}\natexlab{}.
\newblock \showarticletitle{Transfer Learning: A New Promising Techniques}.
\newblock \bibinfo{journal}{\emph{Mesopotamian Journal of Big Data}}  \bibinfo{volume}{2023} (\bibinfo{year}{2023}), \bibinfo{pages}{31--32}.
\newblock


\bibitem[Bansal et~al\mbox{.}(2021)]%
        {bansal2021transfer}
\bibfield{author}{\bibinfo{person}{Monika Bansal}, \bibinfo{person}{Munish Kumar}, \bibinfo{person}{Monika Sachdeva}, {and} \bibinfo{person}{Ajay Mittal}.} \bibinfo{year}{2021}\natexlab{}.
\newblock \showarticletitle{Transfer learning for image classification using VGG19: Caltech-101 image data set}.
\newblock \bibinfo{journal}{\emph{Journal of ambient intelligence and humanized computing}} (\bibinfo{year}{2021}), \bibinfo{pages}{1--12}.
\newblock


\bibitem[Basha and Rajput(2018)]%
        {basha2018evaluating}
\bibfield{author}{\bibinfo{person}{Syed~Muzamil Basha} {and} \bibinfo{person}{Dharmendra~Singh Rajput}.} \bibinfo{year}{2018}\natexlab{}.
\newblock \showarticletitle{Evaluating the Importance of each Feature in Classification task}. In \bibinfo{booktitle}{\emph{2018 8th International Conference on Communication Systems and Network Technologies (CSNT)}}. IEEE, \bibinfo{pages}{151--155}.
\newblock


\bibitem[Bilik and Horak(2022)]%
        {12}
\bibfield{author}{\bibinfo{person}{Simon Bilik} {and} \bibinfo{person}{Karel Horak}.} \bibinfo{year}{2022}\natexlab{}.
\newblock \bibinfo{title}{SIFT and SURF based feature extraction for the anomaly detection}.
\newblock
\newblock
\urldef\tempurl%
\url{https://doi.org/10.48550/ARXIV.2203.13068}
\showDOI{\tempurl}


\bibitem[Bouguettaya et~al\mbox{.}(2022)]%
        {bouguettaya2022deep}
\bibfield{author}{\bibinfo{person}{Abdelmalek Bouguettaya}, \bibinfo{person}{Hafed Zarzour}, \bibinfo{person}{Ahmed Kechida}, {and} \bibinfo{person}{Amine~Mohammed Taberkit}.} \bibinfo{year}{2022}\natexlab{}.
\newblock \showarticletitle{Deep learning techniques to classify agricultural crops through UAV imagery: A review}.
\newblock \bibinfo{journal}{\emph{Neural Computing and Applications}} \bibinfo{volume}{34}, \bibinfo{number}{12} (\bibinfo{year}{2022}), \bibinfo{pages}{9511--9536}.
\newblock


\bibitem[Burri et~al\mbox{.}(2023)]%
        {burri2023exploring}
\bibfield{author}{\bibinfo{person}{Srinivasa~Rao Burri}, \bibinfo{person}{Sachin Ahuja}, \bibinfo{person}{Abhishek Kumar}, {and} \bibinfo{person}{Anupam Baliyan}.} \bibinfo{year}{2023}\natexlab{}.
\newblock \showarticletitle{Exploring the Effectiveness of Optimized Convolutional Neural Network in Transfer Learning for Image Classification: A Practical Approach}. In \bibinfo{booktitle}{\emph{2023 International Conference on Advancement in Computation \& Computer Technologies (InCACCT)}}. IEEE, \bibinfo{pages}{598--602}.
\newblock


\bibitem[Chan et~al\mbox{.}(2023)]%
        {chan2023xai}
\bibfield{author}{\bibinfo{person}{Ayshah Chan}, \bibinfo{person}{Maja Schneider}, {and} \bibinfo{person}{Marco K{\"o}rner}.} \bibinfo{year}{2023}\natexlab{}.
\newblock \showarticletitle{XAI for Early Crop Classification}. In \bibinfo{booktitle}{\emph{IGARSS 2023-2023 IEEE International Geoscience and Remote Sensing Symposium}}. IEEE, \bibinfo{pages}{2657--2660}.
\newblock


\bibitem[Chollet(2017)]%
        {chollet2017xception}
\bibfield{author}{\bibinfo{person}{Fran{\c{c}}ois Chollet}.} \bibinfo{year}{2017}\natexlab{}.
\newblock \showarticletitle{Xception: Deep learning with depthwise separable convolutions}. In \bibinfo{booktitle}{\emph{Proceedings of the IEEE conference on computer vision and pattern recognition}}. \bibinfo{pages}{1251--1258}.
\newblock


\bibitem[Chugh et~al\mbox{.}(2022)]%
        {chugh2022image}
\bibfield{author}{\bibinfo{person}{Himani Chugh}, \bibinfo{person}{Sheifali Gupta}, \bibinfo{person}{Meenu Garg}, \bibinfo{person}{Deepali Gupta}, \bibinfo{person}{Sapna Juneja}, \bibinfo{person}{Hamza Turabieh}, \bibinfo{person}{Yogita Na}, {and} \bibinfo{person}{Zelalem Kiros~Bitsue}.} \bibinfo{year}{2022}\natexlab{}.
\newblock \showarticletitle{Image retrieval using different distance methods and color difference histogram descriptor for human healthcare}.
\newblock \bibinfo{journal}{\emph{Journal of Healthcare Engineering}}  \bibinfo{volume}{2022} (\bibinfo{year}{2022}).
\newblock


\bibitem[Dai and Wu(2023)]%
        {dai2023improved}
\bibfield{author}{\bibinfo{person}{Yong Dai} {and} \bibinfo{person}{Jiaxin Wu}.} \bibinfo{year}{2023}\natexlab{}.
\newblock \showarticletitle{An Improved ORB Feature Extraction Algorithm Based on Enhanced Image and Truncated Adaptive Threshold}.
\newblock \bibinfo{journal}{\emph{IEEE Access}}  \bibinfo{volume}{11} (\bibinfo{year}{2023}), \bibinfo{pages}{32073--32081}.
\newblock


\bibitem[Dharani et~al\mbox{.}(2021)]%
        {dharani2021review}
\bibfield{author}{\bibinfo{person}{MK Dharani}, \bibinfo{person}{R Thamilselvan}, \bibinfo{person}{P Natesan}, \bibinfo{person}{PCD Kalaivaani}, {and} \bibinfo{person}{S Santhoshkumar}.} \bibinfo{year}{2021}\natexlab{}.
\newblock \showarticletitle{Review on crop prediction using deep learning techniques}. In \bibinfo{booktitle}{\emph{Journal of Physics: Conference Series}}, Vol.~\bibinfo{volume}{1767}. IOP Publishing, \bibinfo{pages}{012026}.
\newblock


\bibitem[Gastounioti and Kontos(2020)]%
        {gastounioti2020time}
\bibfield{author}{\bibinfo{person}{Aimilia Gastounioti} {and} \bibinfo{person}{Despina Kontos}.} \bibinfo{year}{2020}\natexlab{}.
\newblock \showarticletitle{Is it time to get rid of black boxes and cultivate trust in AI?}
\newblock \bibinfo{journal}{\emph{Radiology: Artificial Intelligence}} \bibinfo{volume}{2}, \bibinfo{number}{3} (\bibinfo{year}{2020}), \bibinfo{pages}{e200088}.
\newblock


\bibitem[{Gomez Selvaraj} et~al\mbox{.}(2020)]%
        {01}
\bibfield{author}{\bibinfo{person}{Michael {Gomez Selvaraj}}, \bibinfo{person}{Alejandro Vergara}, \bibinfo{person}{Frank Montenegro}, \bibinfo{person}{Henry {Alonso Ruiz}}, \bibinfo{person}{Nancy Safari}, \bibinfo{person}{Dries Raymaekers}, \bibinfo{person}{Walter Ocimati}, \bibinfo{person}{Jules Ntamwira}, \bibinfo{person}{Laurent Tits}, \bibinfo{person}{Aman~Bonaventure Omondi}, {and} \bibinfo{person}{Guy Blomme}.} \bibinfo{year}{2020}\natexlab{}.
\newblock \showarticletitle{Detection of banana plants and their major diseases through aerial images and machine learning methods: A case study in DR Congo and Republic of Benin}.
\newblock \bibinfo{journal}{\emph{ISPRS Journal of Photogrammetry and Remote Sensing}}  \bibinfo{volume}{169} (\bibinfo{year}{2020}), \bibinfo{pages}{110--124}.
\newblock


\bibitem[Guo et~al\mbox{.}(2018)]%
        {guo2018research}
\bibfield{author}{\bibinfo{person}{Feng Guo}, \bibinfo{person}{Jie Yang}, \bibinfo{person}{Yilei Chen}, {and} \bibinfo{person}{Bao Yao}.} \bibinfo{year}{2018}\natexlab{}.
\newblock \showarticletitle{Research on image detection and matching based on SIFT features}. In \bibinfo{booktitle}{\emph{2018 3rd International conference on control and robotics engineering (ICCRE)}}. IEEE, \bibinfo{pages}{130--134}.
\newblock


\bibitem[Gupta et~al\mbox{.}(2019)]%
        {gupta2019improved}
\bibfield{author}{\bibinfo{person}{Surbhi Gupta}, \bibinfo{person}{Munish Kumar}, {and} \bibinfo{person}{Anupam Garg}.} \bibinfo{year}{2019}\natexlab{}.
\newblock \showarticletitle{Improved object recognition results using SIFT and ORB feature detector}.
\newblock \bibinfo{journal}{\emph{Multimedia Tools and Applications}}  \bibinfo{volume}{78} (\bibinfo{year}{2019}), \bibinfo{pages}{34157--34171}.
\newblock


\bibitem[Howard et~al\mbox{.}(2019)]%
        {howard2019searching}
\bibfield{author}{\bibinfo{person}{Andrew Howard}, \bibinfo{person}{Mark Sandler}, \bibinfo{person}{Grace Chu}, \bibinfo{person}{Liang-Chieh Chen}, \bibinfo{person}{Bo Chen}, \bibinfo{person}{Mingxing Tan}, \bibinfo{person}{Weijun Wang}, \bibinfo{person}{Yukun Zhu}, \bibinfo{person}{Ruoming Pang}, \bibinfo{person}{Vijay Vasudevan}, {et~al\mbox{.}}} \bibinfo{year}{2019}\natexlab{}.
\newblock \showarticletitle{Searching for mobilenetv3}. In \bibinfo{booktitle}{\emph{Proceedings of the IEEE/CVF international conference on computer vision}}. \bibinfo{pages}{1314--1324}.
\newblock


\bibitem[Ibrahim et~al\mbox{.}(2020)]%
        {ibrahim2020soft}
\bibfield{author}{\bibinfo{person}{Younis Ibrahim}, \bibinfo{person}{Haibin Wang}, \bibinfo{person}{Man Bai}, \bibinfo{person}{Zhi Liu}, \bibinfo{person}{Jianan Wang}, \bibinfo{person}{Zhiming Yang}, {and} \bibinfo{person}{Zhengming Chen}.} \bibinfo{year}{2020}\natexlab{}.
\newblock \showarticletitle{Soft error resilience of deep residual networks for object recognition}.
\newblock \bibinfo{journal}{\emph{IEEE Access}}  \bibinfo{volume}{8} (\bibinfo{year}{2020}), \bibinfo{pages}{19490--19503}.
\newblock


\bibitem[Imsaengsuk and Pumrin(2021)]%
        {imsaengsuk2021feature}
\bibfield{author}{\bibinfo{person}{Taksaporn Imsaengsuk} {and} \bibinfo{person}{Suree Pumrin}.} \bibinfo{year}{2021}\natexlab{}.
\newblock \showarticletitle{Feature Detection and Description based on ORB Algorithm for FPGA-based Image Processing}. In \bibinfo{booktitle}{\emph{2021 9th International Electrical Engineering Congress (iEECON)}}. IEEE, \bibinfo{pages}{420--423}.
\newblock


\bibitem[Kaur and Sharma(2019)]%
        {kaur2019various}
\bibfield{author}{\bibinfo{person}{Dalvir Kaur} {and} \bibinfo{person}{Sukesha Sharma}.} \bibinfo{year}{2019}\natexlab{}.
\newblock \showarticletitle{Various Feature extraction and classification techniques}. In \bibinfo{booktitle}{\emph{Proceeding of the Second International Conference on Microelectronics, Computing \& Communication Systems (MCCS 2017)}}. Springer, \bibinfo{pages}{633--642}.
\newblock


\bibitem[Liu et~al\mbox{.}(2023)]%
        {liu2023large}
\bibfield{author}{\bibinfo{person}{Mingxuan Liu}, \bibinfo{person}{Subhankar Roy}, \bibinfo{person}{Zhun Zhong}, \bibinfo{person}{Nicu Sebe}, {and} \bibinfo{person}{Elisa Ricci}.} \bibinfo{year}{2023}\natexlab{}.
\newblock \showarticletitle{Large-scale Pre-trained Models are Surprisingly Strong in Incremental Novel Class Discovery}.
\newblock \bibinfo{journal}{\emph{arXiv preprint arXiv:2303.15975}} (\bibinfo{year}{2023}).
\newblock


\bibitem[Lundberg and Lee(2017)]%
        {NIPS2017_8a20a862}
\bibfield{author}{\bibinfo{person}{Scott~M Lundberg} {and} \bibinfo{person}{Su-In Lee}.} \bibinfo{year}{2017}\natexlab{}.
\newblock \showarticletitle{A Unified Approach to Interpreting Model Predictions}. In \bibinfo{booktitle}{\emph{Advances in Neural Information Processing Systems}}, \bibfield{editor}{\bibinfo{person}{I.~Guyon}, \bibinfo{person}{U.~Von Luxburg}, \bibinfo{person}{S.~Bengio}, \bibinfo{person}{H.~Wallach}, \bibinfo{person}{R.~Fergus}, \bibinfo{person}{S.~Vishwanathan}, {and} \bibinfo{person}{R.~Garnett}} (Eds.), Vol.~\bibinfo{volume}{30}. \bibinfo{publisher}{Curran Associates, Inc.}
\newblock
\urldef\tempurl%
\url{https://proceedings.neurips.cc/paper_files/paper/2017/file/8a20a8621978632d76c43dfd28b67767-Paper.pdf}
\showURL{%
\tempurl}


\bibitem[Madake et~al\mbox{.}(2022)]%
        {10119325}
\bibfield{author}{\bibinfo{person}{Jyoti Madake}, \bibinfo{person}{Sanket Shinde}, \bibinfo{person}{Sumitsaurabh Singh}, \bibinfo{person}{Shreyas Talwekar}, \bibinfo{person}{Shripad Bhatlawande}, {and} \bibinfo{person}{Swati Shilaskar}.} \bibinfo{year}{2022}\natexlab{}.
\newblock \showarticletitle{Vision-Based Wilted Plant Detection}. In \bibinfo{booktitle}{\emph{2022 IEEE Conference on Interdisciplinary Approaches in Technology and Management for Social Innovation (IATMSI)}}. \bibinfo{pages}{1--6}.
\newblock
\urldef\tempurl%
\url{https://doi.org/10.1109/IATMSI56455.2022.10119325}
\showDOI{\tempurl}


\bibitem[Mali and Tejaswini(2014)]%
        {mali2014color}
\bibfield{author}{\bibinfo{person}{Sanmukh~N Mali} {and} \bibinfo{person}{ML Tejaswini}.} \bibinfo{year}{2014}\natexlab{}.
\newblock \showarticletitle{Color histogram features for image retrieval systems}.
\newblock \bibinfo{journal}{\emph{International Journal of Innovative Research in Science, Engineering and Technology}} \bibinfo{volume}{3}, \bibinfo{number}{4} (\bibinfo{year}{2014}), \bibinfo{pages}{1094--10946}.
\newblock


\bibitem[Meng et~al\mbox{.}(2023)]%
        {meng2023foundation}
\bibfield{author}{\bibinfo{person}{Fanqing Meng}, \bibinfo{person}{Wenqi Shao}, \bibinfo{person}{Zhanglin Peng}, \bibinfo{person}{Chonghe Jiang}, \bibinfo{person}{Kaipeng Zhang}, \bibinfo{person}{Yu Qiao}, {and} \bibinfo{person}{Ping Luo}.} \bibinfo{year}{2023}\natexlab{}.
\newblock \showarticletitle{Foundation model is efficient multimodal multitask model selector}.
\newblock \bibinfo{journal}{\emph{arXiv preprint arXiv:2308.06262}} (\bibinfo{year}{2023}).
\newblock


\bibitem[Murindanyi et~al\mbox{.}(2023a)]%
        {murindanyi2023interpretable}
\bibfield{author}{\bibinfo{person}{Sudi Murindanyi}, \bibinfo{person}{Ben~Wycliff Mugalu}, \bibinfo{person}{Joyce Nakatumba-Nabende}, {and} \bibinfo{person}{Ggaliwango Marvin}.} \bibinfo{year}{2023}\natexlab{a}.
\newblock \showarticletitle{Interpretable Machine Learning for Predicting Customer Churn in Retail Banking}. In \bibinfo{booktitle}{\emph{2023 7th International Conference on Trends in Electronics and Informatics (ICOEI)}}. IEEE, \bibinfo{pages}{967--974}.
\newblock


\bibitem[Murindanyi et~al\mbox{.}(2023b)]%
        {murindanyi2023explainable}
\bibfield{author}{\bibinfo{person}{Sudi Murindanyi}, \bibinfo{person}{Margaret Nagwovuma}, \bibinfo{person}{Barbara Nansamba}, {and} \bibinfo{person}{Ggaliwango Marvin}.} \bibinfo{year}{2023}\natexlab{b}.
\newblock \showarticletitle{Explainable Ensemble Learning and Trustworthy Open AI for Customer Engagement Prediction in Retail Banking}. In \bibinfo{booktitle}{\emph{Proceedings of the 2023 Fifteenth International Conference on Contemporary Computing}}. \bibinfo{pages}{198--206}.
\newblock


\bibitem[Mwadulo(2016)]%
        {mwadulo2016review}
\bibfield{author}{\bibinfo{person}{Mary~Walowe Mwadulo}.} \bibinfo{year}{2016}\natexlab{}.
\newblock \showarticletitle{A review on feature selection methods for classification tasks}.
\newblock  (\bibinfo{year}{2016}).
\newblock


\bibitem[Obadic et~al\mbox{.}({[n.\,d.]})]%
        {obadic2210exploring}
\bibfield{author}{\bibinfo{person}{I Obadic}, \bibinfo{person}{R Roscher}, \bibinfo{person}{DAB Oliveira}, {and} \bibinfo{person}{XX Zhu}.} \bibinfo{year}{[n.\,d.]}\natexlab{}.
\newblock \showarticletitle{Exploring Self-Attention for Crop-Type Classification Explainability. arXiv 2022}.
\newblock \bibinfo{journal}{\emph{arXiv preprint arXiv:2210.13167}} (\bibinfo{year}{[n.\,d.]}).
\newblock


\bibitem[Oquab et~al\mbox{.}(2023)]%
        {oquab2023dinov2}
\bibfield{author}{\bibinfo{person}{Maxime Oquab}, \bibinfo{person}{Timoth{\'e}e Darcet}, \bibinfo{person}{Th{\'e}o Moutakanni}, \bibinfo{person}{Huy Vo}, \bibinfo{person}{Marc Szafraniec}, \bibinfo{person}{Vasil Khalidov}, \bibinfo{person}{Pierre Fernandez}, \bibinfo{person}{Daniel Haziza}, \bibinfo{person}{Francisco Massa}, \bibinfo{person}{Alaaeldin El-Nouby}, {et~al\mbox{.}}} \bibinfo{year}{2023}\natexlab{}.
\newblock \showarticletitle{Dinov2: Learning robust visual features without supervision}.
\newblock \bibinfo{journal}{\emph{arXiv preprint arXiv:2304.07193}} (\bibinfo{year}{2023}).
\newblock


\bibitem[Patil and Saiyyad(2019)]%
        {patil2019machine}
\bibfield{author}{\bibinfo{person}{Nitin~N Patil} {and} \bibinfo{person}{Mohmmad Ali~M Saiyyad}.} \bibinfo{year}{2019}\natexlab{}.
\newblock \showarticletitle{Machine learning technique for crop recommendation in agriculture sector}.
\newblock \bibinfo{journal}{\emph{Int. J. Eng. Adv. Technol}}  \bibinfo{volume}{9} (\bibinfo{year}{2019}), \bibinfo{pages}{1359--1363}.
\newblock


\bibitem[Peng et~al\mbox{.}(2022)]%
        {peng2022survey}
\bibfield{author}{\bibinfo{person}{Luzhou Peng}, \bibinfo{person}{Bowen Qiang}, {and} \bibinfo{person}{Jiacheng Wu}.} \bibinfo{year}{2022}\natexlab{}.
\newblock \showarticletitle{A survey: Image classification models based on convolutional neural networks}. In \bibinfo{booktitle}{\emph{2022 14th International Conference on Computer Research and Development (ICCRD)}}. IEEE, \bibinfo{pages}{291--298}.
\newblock


\bibitem[Pham et~al\mbox{.}(2017)]%
        {pham2017color}
\bibfield{author}{\bibinfo{person}{Minh-Tan Pham}, \bibinfo{person}{Gr{\'e}goire Mercier}, {and} \bibinfo{person}{Lionel Bombrun}.} \bibinfo{year}{2017}\natexlab{}.
\newblock \showarticletitle{Color texture image retrieval based on local extrema features and Riemannian distance}.
\newblock \bibinfo{journal}{\emph{Journal of Imaging}} \bibinfo{volume}{3}, \bibinfo{number}{4} (\bibinfo{year}{2017}), \bibinfo{pages}{43}.
\newblock


\bibitem[Qian et~al\mbox{.}(2021)]%
        {qian2021mobilenetv3}
\bibfield{author}{\bibinfo{person}{Siying Qian}, \bibinfo{person}{Chenran Ning}, {and} \bibinfo{person}{Yuepeng Hu}.} \bibinfo{year}{2021}\natexlab{}.
\newblock \showarticletitle{MobileNetV3 for image classification}. In \bibinfo{booktitle}{\emph{2021 IEEE 2nd International Conference on Big Data, Artificial Intelligence and Internet of Things Engineering (ICBAIE)}}. IEEE, \bibinfo{pages}{490--497}.
\newblock


\bibitem[Rajpura et~al\mbox{.}(2017)]%
        {rajpura2017transfer}
\bibfield{author}{\bibinfo{person}{Param Rajpura}, \bibinfo{person}{Alakh Aggarwal}, \bibinfo{person}{Manik Goyal}, \bibinfo{person}{Sanchit Gupta}, \bibinfo{person}{Jonti Talukdar}, \bibinfo{person}{Hristo Bojinov}, {and} \bibinfo{person}{Ravi Hegde}.} \bibinfo{year}{2017}\natexlab{}.
\newblock \showarticletitle{Transfer learning by finetuning pretrained CNNs entirely with synthetic images}. In \bibinfo{booktitle}{\emph{National Conference on Computer Vision, Pattern Recognition, Image Processing, and Graphics}}. Springer, \bibinfo{pages}{517--528}.
\newblock


\bibitem[Ravanbakhsh et~al\mbox{.}(2015)]%
        {ravanbakhsh2015action}
\bibfield{author}{\bibinfo{person}{Mahdyar Ravanbakhsh}, \bibinfo{person}{Hossein Mousavi}, \bibinfo{person}{Mohammad Rastegari}, \bibinfo{person}{Vittorio Murino}, {and} \bibinfo{person}{Larry~S Davis}.} \bibinfo{year}{2015}\natexlab{}.
\newblock \showarticletitle{Action recognition with image based CNN features}.
\newblock \bibinfo{journal}{\emph{arXiv preprint arXiv:1512.03980}} (\bibinfo{year}{2015}).
\newblock


\bibitem[Ribeiro et~al\mbox{.}(2016)]%
        {10.1145/2939672.2939778}
\bibfield{author}{\bibinfo{person}{Marco~Tulio Ribeiro}, \bibinfo{person}{Sameer Singh}, {and} \bibinfo{person}{Carlos Guestrin}.} \bibinfo{year}{2016}\natexlab{}.
\newblock \showarticletitle{"Why Should I Trust You?": Explaining the Predictions of Any Classifier}. In \bibinfo{booktitle}{\emph{Proceedings of the 22nd ACM SIGKDD International Conference on Knowledge Discovery and Data Mining}} (San Francisco, California, USA) \emph{(\bibinfo{series}{KDD '16})}. \bibinfo{publisher}{Association for Computing Machinery}, \bibinfo{address}{New York, NY, USA}, \bibinfo{pages}{1135–1144}.
\newblock
\showISBNx{9781450342322}
\urldef\tempurl%
\url{https://doi.org/10.1145/2939672.2939778}
\showDOI{\tempurl}


\bibitem[Roopashree et~al\mbox{.}(2022)]%
        {ROOPASHREE2022111484}
\bibfield{author}{\bibinfo{person}{S. Roopashree}, \bibinfo{person}{J. Anitha}, \bibinfo{person}{T.R. Mahesh}, \bibinfo{person}{V. {Vinoth Kumar}}, \bibinfo{person}{Wattana Viriyasitavat}, {and} \bibinfo{person}{Amandeep Kaur}.} \bibinfo{year}{2022}\natexlab{}.
\newblock \showarticletitle{An IoT based authentication system for therapeutic herbs measured by local descriptors using machine learning approach}.
\newblock \bibinfo{journal}{\emph{Measurement}}  \bibinfo{volume}{200} (\bibinfo{year}{2022}), \bibinfo{pages}{111484}.
\newblock
\showISSN{0263-2241}
\urldef\tempurl%
\url{https://doi.org/10.1016/j.measurement.2022.111484}
\showDOI{\tempurl}


\bibitem[Sada et~al\mbox{.}(2018)]%
        {sada2018histogram}
\bibfield{author}{\bibinfo{person}{Ayumi Sada}, \bibinfo{person}{Yuma Kinoshita}, \bibinfo{person}{Sayaka Shiota}, {and} \bibinfo{person}{Hitoshi Kiya}.} \bibinfo{year}{2018}\natexlab{}.
\newblock \showarticletitle{Histogram-based image pre-processing for machine learning}. In \bibinfo{booktitle}{\emph{2018 IEEE 7th Global Conference on Consumer Electronics (GCCE)}}. IEEE, \bibinfo{pages}{272--275}.
\newblock


\bibitem[Sahann et~al\mbox{.}(2021)]%
        {sahann2021histogram}
\bibfield{author}{\bibinfo{person}{Raphael Sahann}, \bibinfo{person}{Torsten M{\"u}ller}, {and} \bibinfo{person}{Johanna Schmidt}.} \bibinfo{year}{2021}\natexlab{}.
\newblock \showarticletitle{Histogram binning revisited with a focus on human perception}. In \bibinfo{booktitle}{\emph{2021 IEEE Visualization Conference (VIS)}}. IEEE, \bibinfo{pages}{66--70}.
\newblock


\bibitem[Saleem et~al\mbox{.}(2021)]%
        {Saleem2021}
\bibfield{author}{\bibinfo{person}{Muhammad~Hammad Saleem}, \bibinfo{person}{Johan Potgieter}, {and} \bibinfo{person}{Khalid~Mahmood Arif}.} \bibinfo{year}{2021}\natexlab{}.
\newblock \showarticletitle{Automation in Agriculture by Machine and Deep Learning Techniques: A Review of Recent Developments}.
\newblock \bibinfo{journal}{\emph{Precision Agriculture}} \bibinfo{volume}{22}, \bibinfo{number}{6} (\bibinfo{date}{01 Dec} \bibinfo{year}{2021}), \bibinfo{pages}{2053--2091}.
\newblock
\showISSN{1573-1618}
\urldef\tempurl%
\url{https://doi.org/10.1007/s11119-021-09806-x}
\showDOI{\tempurl}


\bibitem[Sanya et~al\mbox{.}(2024)]%
        {SANYA2024109952}
\bibfield{author}{\bibinfo{person}{Rahman Sanya}, \bibinfo{person}{Ann~Lisa Nabiryo}, \bibinfo{person}{Jeremy~Francis Tusubira}, \bibinfo{person}{Sudi Murindanyi}, \bibinfo{person}{Andrew Katumba}, {and} \bibinfo{person}{Joyce Nakatumba-Nabende}.} \bibinfo{year}{2024}\natexlab{}.
\newblock \showarticletitle{Coffee and cashew nut dataset: A dataset for detection, classification, and yield estimation for machine learning applications}.
\newblock \bibinfo{journal}{\emph{Data in Brief}}  \bibinfo{volume}{52} (\bibinfo{year}{2024}), \bibinfo{pages}{109952}.
\newblock
\showISSN{2352-3409}
\urldef\tempurl%
\url{https://doi.org/10.1016/j.dib.2023.109952}
\showDOI{\tempurl}


\bibitem[Selvaraju et~al\mbox{.}(2017)]%
        {Selvaraju_2017_ICCV}
\bibfield{author}{\bibinfo{person}{Ramprasaath~R. Selvaraju}, \bibinfo{person}{Michael Cogswell}, \bibinfo{person}{Abhishek Das}, \bibinfo{person}{Ramakrishna Vedantam}, \bibinfo{person}{Devi Parikh}, {and} \bibinfo{person}{Dhruv Batra}.} \bibinfo{year}{2017}\natexlab{}.
\newblock \showarticletitle{Grad-CAM: Visual Explanations From Deep Networks via Gradient-Based Localization}. In \bibinfo{booktitle}{\emph{Proceedings of the IEEE International Conference on Computer Vision (ICCV)}}.
\newblock


\bibitem[Sergyan(2008)]%
        {sergyan2008color}
\bibfield{author}{\bibinfo{person}{Szabolcs Sergyan}.} \bibinfo{year}{2008}\natexlab{}.
\newblock \showarticletitle{Color histogram features based image classification in content-based image retrieval systems}. In \bibinfo{booktitle}{\emph{2008 6th international symposium on applied machine intelligence and informatics}}. IEEE, \bibinfo{pages}{221--224}.
\newblock


\bibitem[Shamla et~al\mbox{.}(2023)]%
        {shamla2023feature}
\bibfield{author}{\bibinfo{person}{Beevi~A Shamla}, \bibinfo{person}{S Ratheesha}, \bibinfo{person}{Saidalavi Kalady}, {and} \bibinfo{person}{Jenu~James Chakola}.} \bibinfo{year}{2023}\natexlab{}.
\newblock \showarticletitle{Feature Extraction Based on ORB-AKAZE for Echocardiogram View Classification}.
\newblock \bibinfo{journal}{\emph{International journal of electrical and computer engineering systems}} \bibinfo{volume}{14}, \bibinfo{number}{4} (\bibinfo{year}{2023}), \bibinfo{pages}{393--400}.
\newblock


\bibitem[Singh and Singh(2020)]%
        {singh2020image}
\bibfield{author}{\bibinfo{person}{Ankita Singh} {and} \bibinfo{person}{Pawan Singh}.} \bibinfo{year}{2020}\natexlab{}.
\newblock \showarticletitle{Image classification: a survey}.
\newblock \bibinfo{journal}{\emph{Journal of Informatics Electrical and Electronics Engineering (JIEEE)}} \bibinfo{volume}{1}, \bibinfo{number}{2} (\bibinfo{year}{2020}), \bibinfo{pages}{1--9}.
\newblock


\bibitem[Song et~al\mbox{.}(2019)]%
        {song2019object}
\bibfield{author}{\bibinfo{person}{Qian Song}, \bibinfo{person}{Mingtao Xiang}, \bibinfo{person}{Ciara Hovis}, \bibinfo{person}{Qingbo Zhou}, \bibinfo{person}{Miao Lu}, \bibinfo{person}{Huajun Tang}, {and} \bibinfo{person}{Wenbin Wu}.} \bibinfo{year}{2019}\natexlab{}.
\newblock \showarticletitle{Object-based feature selection for crop classification using multi-temporal high-resolution imagery}.
\newblock \bibinfo{journal}{\emph{International Journal of Remote Sensing}} \bibinfo{volume}{40}, \bibinfo{number}{5-6} (\bibinfo{year}{2019}), \bibinfo{pages}{2053--2068}.
\newblock


\bibitem[Sridhar et~al\mbox{.}(2022)]%
        {sridhar2022classification}
\bibfield{author}{\bibinfo{person}{Supraja Sridhar}, \bibinfo{person}{Prabavathy Balasundaram}, {and} \bibinfo{person}{Lekshmi Kalinathan}.} \bibinfo{year}{2022}\natexlab{}.
\newblock \showarticletitle{Classification of Plant Species Using AlexNet Architecture}.
\newblock  (\bibinfo{year}{2022}).
\newblock


\bibitem[Sunil et~al\mbox{.}(2022)]%
        {sunil2022weed}
\bibfield{author}{\bibinfo{person}{GC Sunil}, \bibinfo{person}{Yu Zhang}, \bibinfo{person}{Cengiz Koparan}, \bibinfo{person}{Mohammed~Raju Ahmed}, \bibinfo{person}{Kirk Howatt}, {and} \bibinfo{person}{Xin Sun}.} \bibinfo{year}{2022}\natexlab{}.
\newblock \showarticletitle{Weed and crop species classification using computer vision and deep learning technologies in greenhouse conditions}.
\newblock \bibinfo{journal}{\emph{Journal of Agriculture and Food Research}}  \bibinfo{volume}{9} (\bibinfo{year}{2022}), \bibinfo{pages}{100325}.
\newblock


\bibitem[Tamagnini et~al\mbox{.}(2017)]%
        {tamagnini2017interpreting}
\bibfield{author}{\bibinfo{person}{Paolo Tamagnini}, \bibinfo{person}{Josua Krause}, \bibinfo{person}{Aritra Dasgupta}, {and} \bibinfo{person}{Enrico Bertini}.} \bibinfo{year}{2017}\natexlab{}.
\newblock \showarticletitle{Interpreting black-box classifiers using instance-level visual explanations}. In \bibinfo{booktitle}{\emph{Proceedings of the 2nd workshop on human-in-the-loop data analytics}}. \bibinfo{pages}{1--6}.
\newblock


\bibitem[Tan and Le(2021)]%
        {tan2021efficientnetv2}
\bibfield{author}{\bibinfo{person}{Mingxing Tan} {and} \bibinfo{person}{Quoc Le}.} \bibinfo{year}{2021}\natexlab{}.
\newblock \showarticletitle{Efficientnetv2: Smaller models and faster training}. In \bibinfo{booktitle}{\emph{International conference on machine learning}}. PMLR, \bibinfo{pages}{10096--10106}.
\newblock


\bibitem[Terven and Cordova-Esparza(2023)]%
        {terven2023comprehensive}
\bibfield{author}{\bibinfo{person}{Juan Terven} {and} \bibinfo{person}{Diana Cordova-Esparza}.} \bibinfo{year}{2023}\natexlab{}.
\newblock \showarticletitle{A comprehensive review of YOLO: From YOLOv1 to YOLOv8 and beyond}.
\newblock \bibinfo{journal}{\emph{arXiv preprint arXiv:2304.00501}} (\bibinfo{year}{2023}).
\newblock


\bibitem[Ueki and Kobayashi(2015)]%
        {ueki2015multi}
\bibfield{author}{\bibinfo{person}{Kazuya Ueki} {and} \bibinfo{person}{Tetsunori Kobayashi}.} \bibinfo{year}{2015}\natexlab{}.
\newblock \showarticletitle{Multi-layer feature extractions for image classification—Knowledge from deep CNNs}. In \bibinfo{booktitle}{\emph{2015 International Conference on Systems, Signals and Image Processing (IWSSIP)}}. IEEE, \bibinfo{pages}{9--12}.
\newblock


\bibitem[Vasavi et~al\mbox{.}(2022)]%
        {Vasavi2022}
\bibfield{author}{\bibinfo{person}{Pallepati Vasavi}, \bibinfo{person}{Arumugam Punitha}, {and} \bibinfo{person}{T.~Venkat~Narayana Rao}.} \bibinfo{year}{2022}\natexlab{}.
\newblock \bibinfo{title}{Crop leaf disease detection and classification using machine learning and deep learning algorithms by visual symptoms: A review}.
\newblock , \bibinfo{numpages}{2079-2086}~pages.
\newblock
Issue 2.
\showISSN{20888708}
\urldef\tempurl%
\url{https://doi.org/10.11591/ijece.v12i2.pp2079-2086}
\showDOI{\tempurl}


\bibitem[Wang et~al\mbox{.}(2021)]%
        {wang2021improved}
\bibfield{author}{\bibinfo{person}{Fei Wang}, \bibinfo{person}{Zhendong Liu}, \bibinfo{person}{Hongchun Zhu}, \bibinfo{person}{Pengda Wu}, {and} \bibinfo{person}{Chengming Li}.} \bibinfo{year}{2021}\natexlab{}.
\newblock \showarticletitle{An improved method for stable feature points selection in structure-from-motion considering image semantic and structural characteristics}.
\newblock \bibinfo{journal}{\emph{Sensors}} \bibinfo{volume}{21}, \bibinfo{number}{7} (\bibinfo{year}{2021}), \bibinfo{pages}{2416}.
\newblock


\bibitem[Waqas and Fukushima(2020)]%
        {9306425}
\bibfield{author}{\bibinfo{person}{Muhammad Waqas} {and} \bibinfo{person}{Norishige Fukushima}.} \bibinfo{year}{2020}\natexlab{}.
\newblock \showarticletitle{Comparison of Image Features Descriptions for Diagnosis of Leaf Diseases}. In \bibinfo{booktitle}{\emph{2020 Asia-Pacific Signal and Information Processing Association Annual Summit and Conference (APSIPA ASC)}}. \bibinfo{pages}{934--938}.
\newblock


\bibitem[Yang et~al\mbox{.}(2023)]%
        {yang2023foundation}
\bibfield{author}{\bibinfo{person}{Sherry Yang}, \bibinfo{person}{Ofir Nachum}, \bibinfo{person}{Yilun Du}, \bibinfo{person}{Jason Wei}, \bibinfo{person}{Pieter Abbeel}, {and} \bibinfo{person}{Dale Schuurmans}.} \bibinfo{year}{2023}\natexlab{}.
\newblock \showarticletitle{Foundation models for decision making: Problems, methods, and opportunities}.
\newblock \bibinfo{journal}{\emph{arXiv preprint arXiv:2303.04129}} (\bibinfo{year}{2023}).
\newblock


\bibitem[Yoo et~al\mbox{.}(2022)]%
        {yoo2022extraction}
\bibfield{author}{\bibinfo{person}{Woo~Sik Yoo}, \bibinfo{person}{Kitaek Kang}, \bibinfo{person}{Jung~Gon Kim}, {and} \bibinfo{person}{Yeongsik Yoo}.} \bibinfo{year}{2022}\natexlab{}.
\newblock \showarticletitle{Extraction of Color Information and Visualization of Color Differences between Digital Images through Pixel-by-Pixel Color-Difference Mapping}.
\newblock \bibinfo{journal}{\emph{Heritage}} \bibinfo{volume}{5}, \bibinfo{number}{4} (\bibinfo{year}{2022}), \bibinfo{pages}{3923--3945}.
\newblock


\bibitem[Zhang et~al\mbox{.}(2018a)]%
        {zhang2018residual}
\bibfield{author}{\bibinfo{person}{Xingpeng Zhang}, \bibinfo{person}{Sheng Huang}, \bibinfo{person}{Xiaohong Zhang}, \bibinfo{person}{Wei Wang}, \bibinfo{person}{Qiuli Wang}, {and} \bibinfo{person}{Dan Yang}.} \bibinfo{year}{2018}\natexlab{a}.
\newblock \showarticletitle{Residual inception: a new module combining modified residual with inception to improve network performance}. In \bibinfo{booktitle}{\emph{2018 25th IEEE International Conference on Image Processing (ICIP)}}. IEEE, \bibinfo{pages}{3039--3043}.
\newblock


\bibitem[Zhang et~al\mbox{.}(2018b)]%
        {zhang2018improved}
\bibfield{author}{\bibinfo{person}{Yang Zhang}, \bibinfo{person}{Chao Li}, \bibinfo{person}{Chuqing Cao}, {and} \bibinfo{person}{Yunfeng Gao}.} \bibinfo{year}{2018}\natexlab{b}.
\newblock \showarticletitle{An Improved ORB Feature Point Matching Algorithm}. In \bibinfo{booktitle}{\emph{Proceedings of the 2018 2nd International Conference on Computer Science and Artificial Intelligence}}. \bibinfo{pages}{207--211}.
\newblock


\bibitem[Zhou et~al\mbox{.}(2021)]%
        {zhou2021evaluating}
\bibfield{author}{\bibinfo{person}{Jianlong Zhou}, \bibinfo{person}{Amir~H Gandomi}, \bibinfo{person}{Fang Chen}, {and} \bibinfo{person}{Andreas Holzinger}.} \bibinfo{year}{2021}\natexlab{}.
\newblock \showarticletitle{Evaluating the quality of machine learning explanations: A survey on methods and metrics}.
\newblock \bibinfo{journal}{\emph{Electronics}} \bibinfo{volume}{10}, \bibinfo{number}{5} (\bibinfo{year}{2021}), \bibinfo{pages}{593}.
\newblock


\end{thebibliography}

\end{document}